\documentclass[11pt,fleqn,letterpaper,twoside,reqno,nosumlimits]{amsart}
\usepackage[letterpaper,margin=2.0cm]{geometry}
\usepackage[utf8]{inputenc} 

\usepackage{etoolbox}
\usepackage{amssymb}
\usepackage{multirow}
\usepackage{mathrsfs}  
\usepackage{array} 

\newcommand{\T}{\mathbb{T}}

\patchcmd{\section}{\scshape}{\bfseries}{}{}
\makeatletter
\renewcommand{\@secnumfont}{\bfseries}
\makeatother

\usepackage[dvipsnames]{xcolor}
\patchcmd{\section}{\normalfont}{\normalfont\color{MidnightBlue}}{}{}
\patchcmd{\subsection}{\normalfont}{\normalfont\color{MidnightBlue}}{}{}
\usepackage{pythonhighlight}

\usepackage[toc,page]{appendix}
\usepackage{fancyvrb}
\usepackage{listings}
\usepackage{algorithm}
\usepackage{algpseudocode}
\usepackage{bm}

\usepackage{todonotes}

\usepackage{arydshln} 

\usepackage[normalem]{ulem} 

\usepackage{fancyhdr}
\usepackage{amsmath}

\usepackage{circuitikz}
\usepackage{tikz}
\usetikzlibrary{positioning, calc, arrows.meta,automata,positioning}

\usepackage{overpic}
\usepackage{amsopn}

\newcommand*{\argmin}{\mathrm{arg\,min}}

\newcommand{\F}{\mathcal{F}}
\newcommand{\V}{\mathcal{V}}
\newcommand{\U}{\mathcal{U}}
\newcommand{\W}{\mathcal{W}}

\renewcommand{\hat}{\widehat}

\usepackage{booktabs}
\usepackage{makecell}
\usepackage{caption}
\usepackage{graphicx}
\usepackage{color}
\usepackage{url}
\usepackage{epstopdf}
\usepackage[colorlinks]{hyperref}
\usepackage{comment}
\usepackage{mathabx}
\usepackage{upgreek}
\usepackage{subfig}
\usepackage{parskip}
\usepackage{tabularx} 
\usepackage[capitalize, nameinlink]{cleveref}
\usepackage{yfonts}

\usepackage{accents}

\newtheorem{Theorem}{Theorem}[section]
\newtheorem{Proposition}[Theorem]{Proposition}

\newtheorem{Corollary}[Theorem]{Corollary}
\newtheorem{Remark}[Theorem]{Remark}
\newtheorem{Example}[Theorem]{Example}
\newtheorem{Assumption}[Theorem]{Assumption}

\usepackage{aurical}

\usetikzlibrary{shadows}
\usetikzlibrary{arrows}
\usetikzlibrary{shapes}

\usepackage[numbers]{natbib}

\usepackage[foot]{amsaddr}

\def\<{\big\langle}
\def\>{\big\rangle}

\begin{document}

\title{Operator Learning at Machine Precision}

\author{Aras Bacho\textsuperscript{a,1,2}, Aleksei G. Sorokin\textsuperscript{b,2}, Xianjin Yang\textsuperscript{a,2}, Th\'{e}o Bourdais \textsuperscript{a}, Edoardo Calvello\textsuperscript{a}, Matthieu Darcy\textsuperscript{a}, Alexander Hsu\textsuperscript{c}, Bamdad Hosseini\textsuperscript{c}, Houman Owhadi\textsuperscript{a}}

\footnotetext[1]{To whom correspondence should be addressed. E-mail: bacho@caltech.edu}
\footnotetext[2]{AB, AS, and XY contributed equally to this work.}
\address{\textsuperscript{a} Department of Computing and Mathematical Sciences, California Institute of Technology, Pasadena, CA, USA}
\address{\textsuperscript{b} Department of Applied Mathematics, Illinois Institute of Technology, Chicago, IL, USA}
\address{\textsuperscript{c} Department of Applied Mathematics, University of Washington, Seattle, WA, USA}

\date{}


\begin{abstract}
Neural operator learning methods have garnered significant attention in scientific computing for their ability to approximate infinite-dimensional operators. However, increasing their complexity often fails to substantially improve their accuracy, leaving them on par with much simpler approaches such as kernel methods and more traditional reduced-order models. In this article, we set out to address this shortcoming and introduce CHONKNORIS (Cholesky Newton--Kantorovich Neural Operator Residual Iterative System), an operator learning paradigm that can achieve machine precision. CHONKNORIS draws on numerical analysis: many nonlinear forward and inverse PDE problems are solvable by Newton-type methods. Rather than regressing the solution operator itself, our method regresses the Cholesky factors of the elliptic operator associated with Tikhonov-regularized Newton--Kantorovich updates. The resulting unrolled iteration yields a neural architecture whose machine-precision behavior follows from achieving a contractive map, requiring far lower accuracy than end-to-end approximation of the solution operator. We benchmark CHONKNORIS on a range of nonlinear forward and inverse problems, including a nonlinear elliptic equation, Burgers' equation, a nonlinear Darcy flow problem, the Calder\'{o}n problem, an inverse wave scattering problem, and a problem from seismic imaging. We also present theoretical guarantees for the convergence of CHONKNORIS in terms of the accuracy of the emulated Cholesky factors. Additionally, we introduce a foundation model variant, FONKNORIS (Foundation Newton--Kantorovich Neural Operator Residual Iterative System), which aggregates multiple pre-trained CHONKNORIS experts for diverse PDEs to emulate the solution map of a novel nonlinear PDE. Our FONKNORIS model is able to accurately solve unseen nonlinear PDEs such as the Klein--Gordon and Sine--Gordon equations.
\end{abstract}

\maketitle

\section{Introduction}
Operator learning \cite{boulle2024mathematical,kovachki2024operator, batlle2024kernel} is the problem of approximating, from limited data, an infinite-dimensional mapping $\mathcal{G} :\mathcal{U}\to\mathcal{V}$ between Banach spaces $\mathcal{U}$ and $\mathcal{V}$. Often, $\mathcal{G}$  may be implicitly defined by another operator $\mathcal{F}$ such that $\mathcal{F}(u,\mathcal{G}(u))=0$. Operator learning has attracted significant attention in scientific computing and has become one of the core problems of physics-informed machine learning (PIML) \cite{karniadakis2021physics} for learning equations \cite{brunton2016discovering,rudy2019data,jalalian2025data}, and solving Partial Differential Equations (PDEs) and inverse problems \cite{cuomo2022scientific, chen2021solving, bartolucci2024convolutional, MYES23NIP}. Among 
the most well-known operator learning models are 
Deep Operator Nets (DeepONet) \cite{lu2021learning} and Fourier Neural Operators (FNO) \cite{li2020fourier}. While differing in their parameterizations, operator learning methods are typically designed to learn discretized approximations of operators between function spaces from empirical data.

We consider two main categories of operator learning problems:
\begin{itemize}
    \item \textbf{Category 1}: The target operator is available only through input-output pairs, and the governing equations or boundary conditions are unknown or incomplete.  This often necessitates a purely data-driven approach.
    \item \textbf{Category 2}: The target operator is known to satisfy a set of constraints, typically in the form of a PDE with appropriate boundary conditions.
\end{itemize}

In the context of \textbf{Category 2}, physics-informed operator learning methods
\cite{goswami2023physics} are often used to encode the constraints into an accelerated emulator for the PDE solution. While these approaches can provide substantial acceleration, to the best of our knowledge existing methods cannot match the accuracy of the numerical solvers they emulate.

\subsection{Summary of contributions}\label{sec:contributions}

In this article we focus on operator learning problems under
{\bf Category 2} and make four key contributions:

\begin{itemize}
\item \textbf{Introduction of \textsc{CHONKNORIS}:} We propose a novel neural operator learning method inspired by the Newton--Kantorovich method \cite{polyak2006newton}. The \textsc{CHONKNORIS} model explicitly learns the dependence of the Cholesky factor $\mathcal{R}$ of the
Gauss-Newton Hessian matrix $(\lambda I + \F'(u,v)^*\F(u,v))^{-1}$ on the approximate PDE solution $v$ and the random coefficients $u$.  Integrating \textsc{CHONKNORIS} Cholesky factor predictions $\hat{\mathcal{R}}$ into a quasi-Newton iterative algorithm enables us to emulate PDE solvers to machine precision error.

\item \textbf{Generalization via \textsc{FONKNORIS}:} Building upon \textsc{CHONKNORIS}, we introduce \textsc{FONKNORIS}, a foundation modeling framework which learns the dependence of the Cholesky factors $\mathcal{R}$ on the approximate PDE solution $v$ and a set of coefficients functions $u$ common to a large class of PDEs. The enhanced generalization of \textsc{FONKNORIS} enables machine precision recovery of PDEs not seen during training.

\item \textbf{Enhanced Accuracy and Interpretability:} By explicitly embedding the underlying physics into a kernel interpolation model or a neural network architecture resembling ResNet / Transformer layers, \textsc{CHONKNORIS} significantly surpasses the accuracy limitations of existing operator learning frameworks while providing improved interpretability through the physically meaningful structure of each iteration.

\item \textbf{Theoretical Guarantees.} We establish an inexact Newton--Kantorovich analysis for the \emph{learned} Tikhonov inverse approximate Hessian, yielding rigorous Kantorovich-style convergence guarantees. The theoretical result is instantiated on a nonlinear elliptic PDE, where we explicitly compute the design error for a kernel-based approximation of the learned Cholesky factors.

\item \textbf{Comprehensive Experimental Validation:} Extensive numerical experiments demonstrate that \textsc{CHONKNORIS} and \textsc{FONKNORIS} are capable of machine precision recovery across a wide variety of forward and inverse nonlinear PDE problems. These include a nonlinear elliptic equation, the Burgers' equation, a nonlinear Darcy flow equation, Calder\'on's inverse problem, an inverse wave scattering problem, a seismic imaging full waveform inversion problem, and a \textsc{FONKNORIS} generalization to the Klein--Gordon and Sine--Gordon equations.

\item \textbf{Benchmarking against operator learners.} Compared to strong baselines (kernel methods, Fourier Neural Operators, and Transformer-based Neural Operators), our approach lowers typical relative errors from around $10^{-3}$ or $10^{-2}$ to near \emph{$10^{-16}$} (machine precision) on benchmark forward and inverse problems. 

\end{itemize}

Finally we highlight the practical and scientific use cases of \textsc{CHONKNORIS} and \textsc{FONKNORIS} algorithms: 

\begin{itemize}
    \item \textbf{Known physics.} When the residual map $\mathcal{F}(u,v)=0$ is specified (forward or inverse settings) and high accuracy is required, the solver-emulating updates attain machine precision accuracy in practice, substantially exceeding typical operator-learning baselines.
    \item \textbf{Reliability.} \textsc{CHONKNORIS} is equipped with theoretical guarantees (e.g., convergence and stability under standard assumptions), yielding reproducible and dependable results.
    \item \textbf{Controllable accuracy and cost.} Accuracy is governed by the iteration budget; per-iteration complexity is dominated by Jacobian actions in the form of  two triangular matrix--vector products.
    \item \textbf{Cross-PDE transfer.} Our foundation model, \textsc{FONKNORIS}, learns a single mapping from operator coefficients to Cholesky factors, enabling transfer across PDE families and generalization to previously unseen equations without retraining.
    
\end{itemize}

\subsection{Brief review of the relevant literature}\label{sec:relevant-literature}

 Methods for operator learning can be broadly categorized into three main groups: (1) Artificial neural network (ANN)-based approaches \cite{lu2021learning, li2020fourier, kovachki2023neural}, (2) kernel-based methods \cite{kadri2016operator, nelsen2021random, batlle2024kernel}, and (3) hybrid approaches \cite{owhadi2019kernel, owhadi2023ideas, mora2025operator}.
Kernel-based methods are well-established, offering strong theoretical foundations and convergence guarantees. Theoretical support for ANN-based methods has been 
developed more recently \cite{kovachki2024operator, DeRyck_Mishra_2024, Mishra_Molinaro_2022, reinhardt2024statistical}. ANN methods have 
become popular as they can 
benefit from advanced hardware and software ecosystems, enabling efficient scaling to large datasets. 
When $\mathcal{G}$ corresponds to the solution operator of a PDE, learning $\mathcal{G}$ through evaluations of $\mathcal{F}$ is referred to as physics-informed operator learning \cite{wang2021learning, goswami2023physics, li2024physics}. These methods typically incorporate a loss term to enforce the consistency of the learned operator with the underlying PDE. As mentioned earlier, 
a common issue with operator learning methods is 
their limitations in terms of accuracy. 
For example, in tasks such as mapping diffusion coefficients to solutions in second order elliptic 
PDEs a relative $L^2$ error of around 0.1\% is achieved 
on benchmark data sets, even in the physics-informed setting \cite{kissas2022learning, he2024mgno, li2024physics}.
Adjacent to the above works, the recent papers 
\cite{long2024kernel, jalalian2025data} explored operator learning via equation learning where $\mathcal{F}$ is learned from data and then inverted numerically to estimate $\mathcal{G}$ and achieved significant improvements in accuracy and data efficiency. However, these methods are not true emulators since every evaluation of the learned operator requires the numerical solution of a nonlinear PDE. In parallel, there is a growing interest in \emph{foundation models} for operator learning, where large neural operators are pre-trained across families of PDEs and subsequently adapted to solve new unseen PDEs or tasks, see, e.g., \cite{Bodnar2025Aurora, herde2024poseidon, Ye2024PDEformer}.

These results raise a compelling question: 

\begin{quote}
    \it Can operator learning for PDE problems achieve machine precision or at least approach it, if we incorporate 
    explicit knowledge that the underlying 
    map $\mathcal{G}$ is fully determined by known constraints?
\end{quote}

The recent works \cite{he2024mgno, hao2024newton, FABIANI2025113433, GeSiYa25FNN, GeSiYa25FNFI} 
introduced operator learning techniques based on 
traditional numerical algorithms. In \cite{he2024mgno}, the authors propose a 
multi-grid operator for solving linear PDEs while \cite{hao2024newton, FABIANI2025113433} 
introduce an operator learning algorithm that emulates
the iterative updates of Newton's method for 
solving PDEs. In \cite{GeSiYa25FNN,GeSiYa25FNFI}, the authors introduce Fredholm neural networks, which solve Fredholm integral equations of the second kind by unrolling an iterative fixed-point scheme into a feed-forward architecture. Several PDEs, including the Helmholtz equation, admit such formulations. In \cite{GeSiYa25FNFI}, they report small interior errors and near machine precision accuracy on the boundary for selected benchmarks. By contrast, \cite{FABIANI2025113433} proposes \emph{RandONets}: shallow, one-hidden-layer operator networks that first embed inputs via random projections and then learn only the output weights with linear solvers. For certain linear operators with aligned data, this approach attains near machine precision. The work \cite{hao2024newton} is the closest method to ours. Methodologically, the method in \cite{hao2024newton} learns a neural operator that directly approximates the nonlinear Newton update map and obtains a solution by iterating this learned map, guided by a Newton-informed loss. In contrast, our \textsc{CHONKNORIS} framework regresses Cholesky factors of the underlying linear elliptic operators and hard-wires a regularized Newton--Kantorovich residual iteration as the network architecture. The \textsc{CHONKNORIS} framework therefore does not approximate the full nonlinear Newton step, but only the linear elliptic operator entering the Newton--Kantorovich linearization. Under standard assumptions ensuring that the Newton--Kantorovich iteration based on the exact elliptic operator is contractive, and provided that the learned operator is sufficiently close to this exact operator so as to preserve contraction, increasing the iteration budget systematically reduces the error down to the accuracy level dictated by the underlying numerical discretization.

Gauss--Newton algorithms, along with other quasi-Newton algorithms as well as 
their function space extensions, the Newton--Kantorovich method, have been widely used for solving nonlinear PDEs \cite{quarteroni1994numerical,zeidler2013nonlinear} 
and inverse problems \cite{haber2000optimization, haber2004quasi}. Such algorithms are also central to 
the design of recent RKHS methods for solving PDEs \cite{chen2021solving} as well as 
boosting Physics Informed Neural Net (PINN)-type methods \cite{jnini2024gauss}. The wide applicability and simple 
abstract formulation of quasi-Newton algorithms suggests that 
they can be emulated or approximated using Machine Learning (ML) models, 
further motivating our exposition in light of 
algorithm unrolling ideas in the design of ANN architectures 
\cite{monga2021algorithm,yang2022data}.

Finally, we note that operator learning is deeply related 
to older ideas in scientific computing and applied 
mathematics such as 
computer model emulation \cite{kennedy2001bayesian},
operator compression \cite{feischl2020sparse},
polynomial chaos expansions \cite{xiu2002wiener},
and model order reduction\cite{schilders2008model} among others. 
For more historical remarks in this direction we 
refer the reader to the literature review in \cite{batlle2024kernel}.

\subsection{Outline}

The remainder of this article is organized as follows. \Cref{sec:operator_learning} details the operator learning problem (\Cref{secoplepb}), the Newton--Kantorovich method (\Cref{sec:nk_method}), the Newton--Kantorovich method with Tikhonov regularization (\Cref{sec:nk_method_tikhonov}), the proposed \textsc{CHONKNORIS} method (\Cref{sec:CHONKNORIS}), and the proposed \textsc{FONKNORIS} foundation model (\Cref{sec:FONKNORIS}). \Cref{sec:numerical_experiments} details our numerical experiments including benchmarking \textsc{CHONKNORIS} against existing operator learning methods (\Cref{se:benchmarks}), forward problem modeling with \textsc{CHONKNORIS} (\Cref{sec:forward_problems}), foundation modeling with \textsc{FONKNORIS} (\Cref{sec:foundation_modeling_numerics}), and inverse problem modeling with \textsc{CHONKNORIS} (\Cref{sec:inverse_problems_numerics}). \Cref{sec:theoretical_guarantees} provides theoretical guarantees of the \textsc{CHONKNORIS} method. Finally, \Cref{sec:conclusion_outlook} gives a brief conclusion and outlook of future work.

\section{Operator Learning} \label{sec:operator_learning}

\subsection{The operator learning problem}\label{secoplepb}
Let $\mathcal{U}$ and $\mathcal{V}$ be two separable Banach function spaces. We are interested in learning the operator $\mathcal{G} :\mathcal{U}\to\mathcal{V}$ that is implicitly defined by $\mathcal{F}: \mathcal{U}\times \mathcal{V}\rightarrow \mathcal{W}$:
\begin{equation*}
    \mathcal{G} : u\mapsto \mathcal{G}(u):=v \,\quad\text{such that }\quad \mathcal{F}(u,\mathcal{G}(u))=0 \quad \text{for all }u\in \mathcal{U}.
\end{equation*}
In this work, we assume to know and have full access to the operator $\mathcal{F}$, and thus to its Fr\'{e}chet derivative $\frac{\delta \mathcal{F}}{\delta v}: \mathcal{U}\times \mathcal{V}\rightarrow \mathcal{L}(\mathcal{V},\mathcal{W})$ as well as its adjoint $\frac{\delta \mathcal{F}}{\delta v}^*: \mathcal{U}\times \mathcal{V}\rightarrow \mathcal{L}(\mathcal{W},\mathcal{V})$.

\subsubsection{Nonlinear elliptic PDE as a running example}
As a running example, we consider the following nonlinear elliptic equation supplemented with periodic boundary conditions. We will use this 
PDE throughout the paper to ground our ideas and provide intuition 
for our abstract derivations:
\begin{align}\label{eq:example_elliptic_pde}
\begin{cases}
    - \Delta v(x) + \kappa v(x)^3=u(x)&\text{ for } x\in (0,1)\\
    v(0)=v(1)
\end{cases}.
\end{align}
The corresponding differential operator and its derivative are
\begin{equation} \label{eq:elliptic_pde_calF}
    \mathcal{F}(u,v)=- \Delta v + \kappa v^3-u \qquad\text{and}\quad \left[\frac{\delta \mathcal{F}}{\delta v}(u,v)\right](h) = [-\Delta + 3\kappa v^2](h)
\end{equation}
respectively. Our task is to learn an approximation of $\mathcal{G}$ such that for $v=\mathcal{G}(u)$ the above equation is satisfied in some suitable sense, e.g., in the classical or weak  sense. In other words, we aim to learn the solution operator that maps the right hand side $u$ to the solution $v$ of \eqref{eq:example_elliptic_pde}.

A suitable analytical framework may set $\mathcal{U}=\mathcal{W}=\mathrm{C}_{per}(\mathbb{T})$, the space of periodic continuous functions on $\mathbb{T}$, and either $\mathcal{V}=\mathrm{C}_{per}^2(\mathbb{T})$, the space of periodic twice differentiable functions on $\Omega$ for classical solutions, or $\V:=H^1_{\mathrm{per},0}(\T)=\Bigl\{u\in H^1_{\mathrm{per}}(\T):\int_0^1 u=0\Bigr\}$ for weak solutions where $W:=V^*\simeq H^{-1}_{\mathrm{per},0}(\T)$
with $\|u\|_V:=\|u'\|_{L^2(0,1)}$.

\subsection{The Newton--Kantorovich method} \label{sec:nk_method}

The Newton--Kantorovich (NK) method \cite{polyak2006newton} is a generalization of Newton's method to Banach spaces. First, we outline the method with the nonlinear elliptic PDE defined in \eqref{eq:example_elliptic_pde} and \eqref{eq:elliptic_pde_calF}. To solve this equation, we start from an initial guess $v_0$, and iteratively perturb the approximation using successive linearizations around each estimate. Writing $v_n$ for the current estimate and $v_{n+1}:=v_n+\delta v_n$ for the next estimate, we have:
\begin{align*}
    0 &= -\Delta (v_n+\delta v_n)+\kappa (v_n+\delta v_n)^3 - u\\
    &=\left[-\Delta (v_n)+\kappa (v_n)^3 - u\right] + (-\Delta \delta v_n +3\kappa v_n^2\delta v_n)+o(\delta v_n)\\
    &=\mathcal{F}(u,v_n) + (-\Delta + 3 \kappa v_n^2)\delta v_n +o(\delta v_n).
\end{align*}
Ignoring higher orders of approximation, this defines the update $\delta v_n$ as the solution of a $v_n$-dependent linear PDE. Our critical insight  is that learning to solve this linearized PDE is much simpler than directly solving the original nonlinear PDE. 
This methodology can be generalized to any operator equation $\mathcal{F}(u,v)=0$: We have \begin{equation} \label{eq:linear}
    0=\mathcal{F}(u,v_n+\delta v_n)\approx \mathcal{F}(u,v_n)+\left[\frac{\delta \mathcal{F}}{\delta v}(u,v_n)\right](\delta v_n)
\end{equation} 
where again $\frac{\delta \mathcal{F}}{\delta v}$ is the Fr\'{e}chet derivative of $\mathcal{F}$. The NK method proceeds to find $\delta v_n$ by solving the linear problem in \eqref{eq:linear} giving
\begin{equation}\label{eq:NK.increment}
\delta v_n= -\left( \frac{\delta \mathcal{F}}{\delta v}(u,v_n) \right)^{-1} \mathcal{F}(u,v_n). 
\end{equation}
Therefore to emulate the NK flow, it is sufficient to learn the term $\left(\frac{\delta \mathcal{F}}{\delta v}(u,v_n) \right)^{-1}$.
\subsection{The Newton--Kantorovich method with Tikhonov regularization} \label{sec:nk_method_tikhonov}
Limited data and the potential ill-conditioning of the operator $\frac{\delta \mathcal{F}}{\delta v}$ can lead to inaccuracies and 
instabilities if we try to directly use the iteration in \eqref{eq:NK.increment}. To address this issue, it is necessary to regularize the problem. We therefore consider a Tikhonov-regularized variant of the NK method by defining
\begin{equation} \label{eq:minimization}
    \delta v_n = \argmin_{\delta v\in \V}\left \lbrace \Big \|\mathcal{F}(u,v_n)+\frac{\delta \mathcal{F}}{\delta v}(u,v_n)[\delta v] \Big \|_{\mathcal{W}}^2+\lambda \|\delta v\|_{\mathcal{V}}^2\right \rbrace
\end{equation}
where $\lambda>0$ is the regularization parameter that balances robustness and accuracy; larger values of $\lambda$ slow down convergence and improve robustness while smaller values lead to faster algorithms that are less stable.
Assuming that the underlying spaces are Hilbert, the first-order optimality
condition for \eqref{eq:minimization} yields a (linear) normal equations that can be solved explicitly. This results in the following iterative scheme
\begin{align}
\delta v_n
&= -\Big(\big(\tfrac{\delta \mathcal F}{\delta v}(u,v_n)\big)^{\!*}
      \tfrac{\delta \mathcal F}{\delta v}(u,v_n)+\lambda I\Big)^{-1}
      \big(\tfrac{\delta \mathcal F}{\delta v}(u,v_n)\big)^{\!*}\,
      \mathcal F(u,v_n),\label{eq:RegNK.normal}\\
v_{n+1} &= v_n+\delta v_n,
\label{eq:RegNK.increment}
\end{align} where $\frac{\delta \mathcal{F}}{\delta v}^*$ denotes the Hilbert adjoint of the Fr\'{e}chet derivative of $\mathcal{F}$.
This is precisely the Hilbert-space analogue of the Levenberg--Marquardt step \cite{nocedal2006numerical}. We note that this iterative scheme leads to a connection with the attention mechanism from transformers \cite{vaswani2017attention, Calvello2024Continuum} that we highlight in \cref{se:attention}, leaving further investigation to future work.

In Banach spaces, a (unique) minimizer of \eqref{eq:minimization} still exists
under standard assumptions (e.g., uniform convexity/smoothness and suitable
conditions on \(\mathcal F\)). The corresponding optimality conditions replaces inner products by duality pairings and involves the duality mappings \(J_V\) and \(J_W\), leading to a nonlinear equation of the form
\[
\big(\tfrac{\delta \mathcal F}{\delta v}(u,v_n)\big)^*
J_W\!\big(\mathcal F(u,v_n)+\tfrac{\delta \mathcal F}{\delta v}(u,v_n)\,
\delta v_n\big)
+ \lambda\, J_V(\delta v_n)=0,
\]
which can be solved by standard monotone-operator or Newton-type methods.
For clarity, we confine our analysis to the Hilbert-space setting.

Since we assumed that the adjoint operator $\frac{\delta \mathcal{F}}{\delta v}^*$ is known, the most significant computational burden in computing the update $\delta v_n$ is, in general, to compute $Q(u,v_n)=\left(\left(\frac{\delta \mathcal{F}}{\delta v}(u,v_n)\right)^* \frac{\delta \mathcal{F}}{\delta v}(u,v_n) + \lambda I\right)^{-1}$. For that reason, we only learn the parametric operator $(u,v)\mapsto Q(u,v)$. This operator is symmetric, positive definite and linear, but it depends nonlinearly on both $u$ and $v_n$. The next remark justifies learning \(Q\), as it can be done efficiently.

\begin{Remark}[Reduction to parametric elliptic operator learning] \label{re:reduction_to_elliptic}
Let us write $\mathcal{L}_+(\mathcal{V},\mathcal{V})$ for the space of symmetric positive linear operators from $\mathcal{V}$ to $\mathcal{V}$. We notice that $Q$ is indeed a mapping from $\mathcal{U}\times \mathcal{V}$ to $\mathcal{L}_+(\mathcal{V},\mathcal{V})$. Then, assuming that the adjoint operator $\frac{\delta \mathcal{F}}{\delta v}^*$ is known, we can compute $\mathcal{Q}\frac{\delta \mathcal{F}}{\delta v}^*.$
The operator $\mathcal{Q}^{-1}$ is indeed an elliptic differential 
operator whose solution map $\mathcal{Q}$ can be learned efficiently 
\cite{SchOwh24SRES,ChScDe24LFMF}.
\end{Remark}

The subsequent remark shows that our analogue of the Levenberg--Marquardt algorithm, as in the finite-dimensional case, interpolates between gradient descent and the Newton--Kantorovich method.

\begin{Remark}[Interpolation between Gradient Descent and NK]\label{remark:lambda-limits}
In the Hilbert space setting, we can formally write 
\begin{equation}\label{eq:approx}
\left\|\mathcal{F}(u,v_n)+\left[\frac{\delta \mathcal{F}}{\delta v}(u,v_n)\right](\delta v)\right\|_{\mathcal{W}}^2
= \|\mathcal{F}(u,v_n)\|_{\mathcal{W}}^2
+ 2\left\langle \delta v, \left[\frac{\delta \mathcal{F}}{\delta v}(u,v_n)\right]^*\mathcal{F}(u,v_n)\right\rangle_{\mathcal{V}}
+\mathcal{O}(\|\delta v\|^2_{\mathcal{V}}), 
\end{equation}
and see that $\delta v_n$ given by \eqref{eq:RegNK.normal} approximates the NK increment \eqref{eq:NK.increment} as $\lambda \downarrow 0$ 
while it is proportional to a gradient descent update for the left-hand side of \eqref{eq:approx} as $\lambda \uparrow  \infty$.  Therefore, the regularization parameter $\lambda$ allows us to interpolate
between the NK algorithm and gradient descent, enabling a controlled balance between convergence stability and update accuracy.
\end{Remark}

\subsection{\textsc{CHONKNORIS}} \label{sec:CHONKNORIS}

We are now ready to introduce \textsc{CHONKNORIS} based on the idea of using operator learning to efficiently learn the solution operator $\mathcal{Q}$ as mentioned in \Cref{re:reduction_to_elliptic}. More precisely, \textsc{CHONKNORIS} uses the same update rule as in \eqref{eq:RegNK.increment}, but replaces the costly operator inversion in the definition of $\mathcal{Q}$ with a learned approximation. In practice, we work with an arbitrary discretization of input and output spaces $\mathcal{U}$, $\mathcal{V}$, and $ \mathcal{W}$. This discretization can be derived from various numerical methods such as finite-elements, finite-differences, spectral methods, or any other discretization method. Thus, choosing an appropriate parametrization of our input and output functions, \textsc{CHONKNORIS} can be made discretization-invariant following the optimal recovery approach in \cite{batlle2024kernel} or the Fourier neural operator (FNO) framework in \cite{li2020fourier}.  Consequently, $\mathcal{Q}(u,v_n)$ is discretized as a positive definite matrix. 
We choose to approximate the Cholesky factors of $\mathcal{Q}$ to enforce positivity, which stabilizes learning, guarantees descent directions, reduces the number of learned parameters, and allows for efficient triangular solves. We train a surrogate operator $\hat{\mathcal{R}}$,
parameterized for instance as a neural operator or a kernel-based model, such that
\begin{align}\label{eq:cholesky_approximation}
    &\hat{\mathcal{R}}(u,v)\hat{\mathcal{R}}(u,v)^T\approx \mathcal{Q}(u,v) = \left[\left(\frac{\delta \mathcal{F}}{\delta v}(u,v)\right)^* \left(\frac{\delta \mathcal{F}}{\delta v}(u,v)\right) + \lambda I\right]^{-1}\\
    &\hat{\mathcal{R}}(u,v) \text{ is upper triangular.}
\end{align}

To that end, we use a traditional NK solver to generate training data for our method. Using input data $u^{(1)},\dots,u^{(M)} \sim \mu$ that are sampled  from a probability measure $\mu$ supported on $\mathcal{U}$, we run the NK solver for $n_{warm}\in \mathbb{N}$ steps to generate flow data $v_k^{(m)}$ where $m=1,\dots, M$ and $k=0,\dots n_{warm}$. This flow data lives along the \emph{true} NK method trajectory to mitigate the curse of dimensionality and avoid generating data that is not seen in the NK iteration scheme. We can then compute the Cholesky factors of $\mathcal{Q}(u^{(m)}, v_k^{(m)})$ and train $\hat{\mathcal{R}}$ to approximate them using a the usual regularized mean squared error loss. Note that it is also possible to vary the Tikhonov regularization $\lambda$, introducing an additional dependency for $\hat{\mathcal{R}}$, i.e., $\hat{\mathcal{R}}(u,v,\lambda)$. This is particularly useful for ill-posed problems, such as inverse problems, to speed up the convergence and obtaining higher accuracy. 

\begin{figure}[htp]
\centering
\begin{tikzpicture}[>=Stealth,
    node distance=1.25cm,
    io/.style   = {draw,circle,fill=SkyBlue!60,minimum size=1.5 em,font=\small},
    aux/.style  = {draw,circle,fill=Orange!40,minimum size=1.5em,font=\small},
    block/.style= {draw,rounded corners,fill=YellowGreen!25,
                   minimum height=1.6em,minimum width=2.9cm,font=\small},
    block2/.style= {draw,rounded corners,fill=Yellow!25,
                   minimum height=1.6em,minimum width=2.9cm,font=\small},               
    plus/.style = {draw,circle,fill=RedOrange!55,inner sep=0pt,
                   minimum size=1.35em,font=\small},
    final/.style = {draw,circle,fill=SkyBlue!60,inner sep=0pt,
                   minimum size=1.35em,font=\small},
    dashed box/.style = {draw,dashed,rounded corners}
]

\begin{scope}[yshift=-1 em, xshift=-2cm, scale=0.85, transform shape]

\node[io] (v1) {$v_0$};
\node[block, right=0.5 cm of v1] (F1) {Iteration 1};
\node[io, right=0.5 cm of F1] (v2) {$v_1$};
\node[block, right=0.5cm of v2] (F2) {Iteration 2};
\node[io, right=0.3cm of F2] (v3) {$v_2$};
\node[right=0.3cm of v3] (dots) {$\bullet\;\bullet\;\bullet$};
\node[block, right=0.3cm of dots] (FT) {Iteration $T$};
\node[io, right=0.5 cm of FT] (vT) {$v_T$};

\draw[->] (v1) -- (F1);
\draw[->] (F1) -- (v2);
\draw[->] (v2) -- (F2);
\draw[->] (F2) -- (v3);
\draw[->] (v3) -- (dots);
\draw[->] (dots) -- (FT);
\draw[->] (FT) -- (vT);

\draw[dashed] (F2.south) ++(0,-0.2) -- ++(-7.1,-1.5);
\draw[dashed] (F2.south) ++(0,-0.2) -- ++(8.5,-1.5);

\end{scope}

\begin{scope}[yshift=-3.2cm, xshift=1cm] 
  \draw[box] (-3.3,1.1) rectangle (10.3,-2.1);

  \node[block2, minimum height=1.3cm, minimum width=7.9cm] (core) at (2.4,0.2) {};

  \node[] at (2.4,0.2)   (R) {$\delta v = -\widehat{\mathcal{R}}(u,v_n) \widehat{\mathcal{R}}^T(u,v_n) \left[ \frac{\delta \mathcal{F}}{\delta v}(u,v_n)\right]^*
\mathcal{F}(u, v_n)$};

    \node[left=1.0cm of core.west] (dummy) {};
  \node[io, below=0.4cm of dummy] (vin) {$v_n$};
  
  \node[below=0.2cm of vin] (dummy2) {};

  \node[aux, below=1 cm of R] (W) {$\mathrm{id}$};
  
  \node[plus, right=10 cm of vin] (add) {$+$};
  
  \node[io, right=0.5 cm of add] (sigma) {$v_{n+1}$};

   \node[below=0.9 cm of add] (dummy3) {};

  \draw[->] (vin) to[out=20,in=180] (R);
  \draw[->] (R) to[out=0,in=180, looseness=0.7] (add);

  \draw[->] (vin) to[out=-20,in=180] (W);
  \draw[->] (W) to[out=0,in= 200] (add);
  \draw[->] (add) -- (sigma);

\end{scope}

\node[anchor=west, font=\small] at (-2.9,-1.6) {(b)};
\node[anchor=west, font=\small] at (-2.9,0.6) {(a)};

\end{tikzpicture}
\caption{
\textsc{CHONKNORIS}.  
(a) An initial guess \(v_0\) for the true solution \(v = \mathcal{G}(u)\) is iteratively refined by adding a correction term. 
(b) Each iteration consists of two steps:  
First, compute the correction term 
$\delta v = -\big(\widehat{\mathcal{R}}\widehat{\mathcal{R}}^* \big[{\tfrac{\delta \mathcal{F}}{\delta v}}\big]^*\mathcal{F}\big)(u,v_k),$
where \(\widehat{\mathcal{R}}\) is a learned surrogate for the Cholesky factors of 
\((\tfrac{\delta \mathcal{F}}{\delta v}\tfrac{\delta \mathcal{F}}{\delta v}^{T}+\lambda I)^{-1}\),  \(\mathcal{F}\) is the forward map with its Fr\'{e}chet derivative \(\tfrac{\delta \mathcal{F}}{\delta v}\), $v_k$ is the current approximation of the desired function $v=G(u)$, and $u$ is the input for which we seek the solution. 
Next, update the current approximation via \(v_{k+1} = v_k + \delta v\). 
}
\label{fig:tikz_fno}
\end{figure}
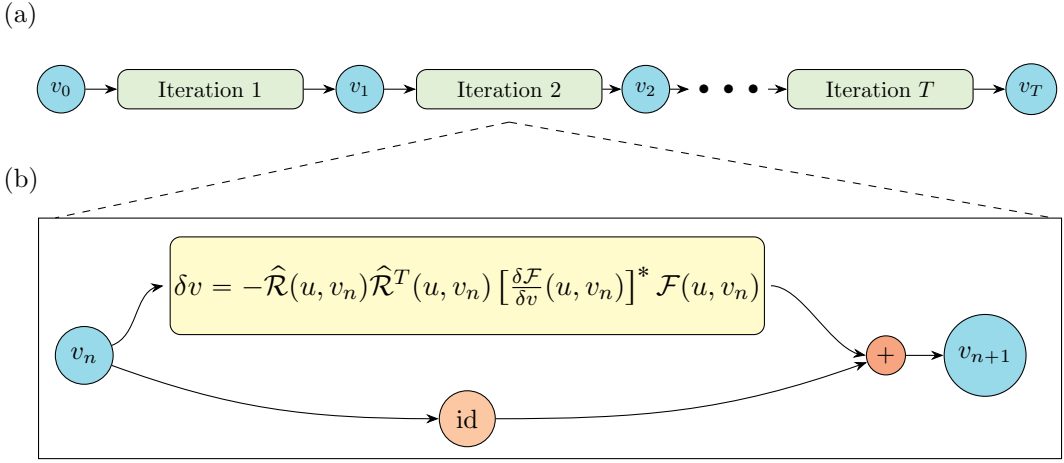

Once we have trained $\hat{\mathcal{R}}$, the \textsc{CHONKNORIS} approximation is $\hat{\mathcal{G}}:u\mapsto \hat{v}_N$, where 
\begin{equation}\label{chunky-update}
    \hat{v}_{n+1}=\hat{v}_n-\alpha_n\hat{\mathcal{R}}(u,\hat{v}_n,\lambda_n)\hat{\mathcal{R}}(u,\hat{v}_n,\lambda_n)^T\left(\frac{\delta \mathcal{F}}{\delta v} (u,\hat{v}_n)\right)^* \mathcal{F}(u,\hat{v}_n), \quad n =0, \dots, N-1.
\end{equation} For a sufficiently large number or inference iterations $N$, we expect $\hat{\mathcal{G}}(u)= \hat{v}_N\approx \mathcal{G}(u)$. Here, $\alpha_n$ is a learning rate schedule and $\lambda_n$ the Tikhonov regularization schedule. Both schedules are chosen using line-search. 
Thus, \textsc{CHONKNORIS} is an emulator of the \emph{true} NK method.
We summarize the resulting algorithm in \Cref{alg:UnifiedNORIS} and 
give a visual depiction of the post-training operator in \Cref{fig:tikz_fno}.

\begin{algorithm}
\caption{\textsc{CHONKNORIS}/\textsc{FONKNORIS}}
\label{alg:UnifiedNORIS}
\begin{algorithmic}[1]
\State \textbf{Inputs:} training data size $M$; residual map $\mathcal{F}$; Jacobian $J(u,v)=\frac{\delta \mathcal{F}}{\delta v}(u,v)$; adjoint $J^*$; measure $\mu$; NK warm-up steps $n_{\text{warm}}$; model $\hat{\mathcal{R}}_\theta$ (lower-triangular, $\mathrm{diag}>0$); flow relaxation $\lambda_\mathrm{flow}$; training relaxation set $\lambda_{\text{train}}$
\State \textbf{Mode:} choose \textit{parameterization} $\Phi$ and \textit{Jacobian builder} $\mathsf{BuildJ}$:
\begin{itemize}\setlength\itemsep{0pt}
    \item \textsc{CHONKNORIS}: $\Phi(u,v)=(u,v)$,\; $\mathsf{BuildJ}(u,v)=J(u,v)$
    \item \textsc{FONKNORIS}: $\Phi(u,v)=(a(u,v),b(u,v),c(u,v))$,\; $\mathsf{BuildJ}(a,b,c)=a\,\partial_{xx}+b\,\partial_x+c$
\end{itemize}

\vspace{.25em}
\State \textbf{Data (offline):}
\For{$m=1,\dots,M$}
  \State sample $u^{(m)}\sim\mu$
  \State initialize $v_0$
  \For{$i=0,\dots,n_{\text{warm}}$} \Comment{solver warmup}
    \State $z_i \gets \Phi(u^{(m)},v_i)$ \Comment{$z_i=(u^{(m)},v_i)$ in \textsc{CHONKNORIS}, or $z_i=(a,b,c)$ in \textsc{FONKNORIS}}
    \State $J_i \gets \mathsf{BuildJ}(z_i)$ 
    \State $v_{i+1}\gets v_i-\left(J_i^* J_i+\lambda_\mathrm{flow}I\right)^{-1}J_i^* \mathcal{F}(u^{(m)},v_i)$
    \For{$\lambda\in\lambda_{\text{train}}$}
        \State $R=\left(\mathrm{chol}_{\text{lower}}(J_i^* J_i + \lambda I)\right)^{-*}$ \Comment{$R {R}^*=(J_i^* J_i + \lambda I)^{-1}$}
        \State store training data $\big(z_i,\lambda,\,R\big)$
    \EndFor
  \EndFor
\EndFor

\vspace{.25em}
\State \textbf{Learn (offline):}\quad $\displaystyle \min_{\theta}\;\sum \big\|\hat{\mathcal{R}}_\theta(z,\lambda)-R\big\|_F^2$;\ \ enforce lower-triangular with positive diagonal.

\vspace{.25em}
\State \textbf{Evaluate (online):} given $u$, $v_0$, budget $N$; initial values $\alpha_\mathrm{predict}/\lambda_{\text{train}}$.
\For{$n=0,\dots,N-1$}
  \State $r\gets \mathcal{F}(u,v_n)$
  \State $z_n \gets \Phi(u,v_n)$
  \State $J_n \gets \mathsf{BuildJ}(z_n)$
  \State $R\gets \hat{\mathcal{R}}_\theta(z_n,\lambda_n)$
  \State $\delta v\gets -\alpha_n R^* R J_n^* r$
  \State $v_{n+1}\gets v_n+\delta v$
  \State choose $\alpha_n,\lambda_n$ by repeating the above steps until $\lVert \mathcal{F}(u,v_{n+1}) \rVert \ll \lVert r \rVert$ \Comment{e.g., using line search}
  \State \textbf{Stop} if $\|r\|$ and/or $\|\delta v\|$ below given tolerances
\EndFor
\State \textbf{Return } $\hat v = v_N$
\end{algorithmic}
\end{algorithm}

\subsection{\textsc{FONKNORIS}} \label{sec:FONKNORIS}
In this section, we introduce \textsc{FONKNORIS}, a foundational model variant of \textsc{CHONKNORIS}. We illustrate this method using our running  example of a nonlinear elliptic PDE. First, we notice that in 1D the Fr\'{e}chet derivative of the nonlinear elliptic PDE in \eqref{eq:elliptic_pde_calF}, $\frac{\delta \mathcal{F}}{\delta v}(u,v)=(-\partial_{xx} + 3 \kappa v^2)$, can be expressed as
\begin{equation} \label{eq:Fonknoris}
    \left[\frac{\delta \mathcal{F}}{\delta v}(u,v)\right](h)(x) = \left[a(u,v) \partial_{xx}  + b(u,v) \partial_x  + c(u,v)\right]h(x),
\end{equation}
for coefficient functions $a=-1$, $b=0$, $c=3\kappa v^2$. While \textsc{CHONKNORIS} learns to predict the Cholesky factors of the Tikhonov-regularized inverse of this operator as a function of $v_n$ and $u$, i.e., $\hat{\mathcal{R}}(u, \hat{v}_n)$, \textsc{FONKNORIS} aims to learn the same Cholesky factors as a function of the coefficient functions $a,b,$ and $c$, i.e., $\hat{\mathcal{R}}(a,b,c)$. This implies that, once the model $\hat{\mathcal{R}}(a,b,c)$ is trained on a sufficiently large and diverse dataset of $(a,b,c)$ combinations, we are able to predict the Cholesky factors arising from any PDE whose Fr\'{e}chet derivative is of the form \eqref{eq:Fonknoris}, i.e., any local PDE of second order. We note that the coefficient functions $a,b,$ and $c$ depend on and are uniquely determined by the operator $\mathcal{F}$. 

To this end, \textsc{FONKNORIS} seeks to train a data-driven surrogate operator $\hat{\mathcal{R}}$, such as a neural operator or a kernel-based model, where
\begin{align*}
    &\hat{\mathcal{R}}(a,b,c)\hat{\mathcal{R}}(a,b,c)^T\approx \left(\left[a \partial_{xx}  + b \partial_x  + c\right]\left[a \partial_{xx}  + b \partial_x  + c\right]^* + \lambda I\right)^{-1}, \\
    &\hat{\mathcal{R}}(a,b,c)\text{ is lower triangular}.
\end{align*}
This model is \emph{trained once, and generalizes to any equation} with a Jacobian of the form \eqref{eq:Fonknoris}. Then, the \textsc{FONKNORIS} approximation is $\hat{\mathcal{G}}:u\mapsto \hat{v}_N$, where 
\begin{align*}
    \hat{v}_{n+1}=\hat{v}_n-\alpha_n\hat{\mathcal{R}}(a_n,b_n,c_n)\hat{\mathcal{R}}(a_n,b_n,c_n)^T\left[a_n \partial_{xx}  + b_n \partial_x  + c_n\right]^*\mathcal{F}(u,\hat{v}_n)\\
    \text{where }a_n=a(u,v_n),\ b_n=b(u,v_n),\ c_n=c(u,v_n).
\end{align*} Again, for a sufficiently large number or iterations $N$, we expect $\hat{\mathcal{G}}(u)= \hat{v}_N\approx \mathcal{G}(u)$. Note that for \textsc{FONKNORIS}, we add an intermediate step that, given $u$ and $v_n$, computes the coefficient functions $a_n=a(u,v_n),\ b_n=b(u,v_n),\ c_n=c(u,v_n)$ and passes them to the model $\hat{\mathcal{R}}$, hence 
we assume knowledge of these functions.

Furthermore, we note that this 1D example can be easily extended to any dimension and to any nonlinear operator. For local differential operators of order $k\in \mathbb{N}$, we observe that 
\begin{align*}\label{eq: FONK}  
    \frac{\partial \mathcal{F}}{\partial v}(u,v) = \sum_{\vert \alpha\vert \leq k} \beta^\alpha(u,v) D^\alpha 
\end{align*}
where \( \alpha \in \mathbb{N}^k \) is a multi-index, \( D^\alpha \) denotes a linear differential operator, and $\beta^\alpha$ denotes the coefficient functions. For nonlinear operators that are nonlocal, the Fr\'{e}chet derivative also contains linear nonlocal operators. However, the principle remains the same: The Fr\'{e}chet derivative can still be parameterized in terms of the coefficient functions. \textsc{FONKNORIS} can be regarded as a \emph{foundation model} for local differential equations, as it is trained once on a diverse collection of PDEs through their coefficient functions $\beta^\alpha$ and generalizes to unseen equations without retraining. By learning a universal mapping from operator coefficients to inverse operators, \textsc{FONKNORIS} provides a reusable model that captures the shared structure underlying  broad classes of  PDEs. The \textsc{FONKNORIS} algorithm is summarized in \Cref{alg:UnifiedNORIS}.

\section{Numerical Experiments} \label{sec:numerical_experiments}

In this section we present various numerical experiments
that verify the ability of \textsc{CHONKNORIS} and \textsc{FONKNORIS}  in 
emulating various PDE and inverse problem solution maps to machine 
precision.

\subsection{Benchmarking Summary}\label{se:benchmarks}

To assess the performance of the CHONKORIS method, we benchmark against the kernel/Gaussian process (GP) operator learning framework of \cite{batlle2024kernel}, Fourier neural operators (FNOs) \cite{li2020fourier}, and transformer neural operator (TNO) \cite{Calvello2024Continuum}. We deploy these three models as purely data-driven baselines which learn the operator $\mathcal{G}$ directly from input-output pairs $(u,v)$. Note that $v$ is given by the last iterate of the Newton--Kantorovich method, so the baseline methods do not have access to the intermediate steps of the solver. These baselines are not intended as a comprehensive benchmark; rather, they serve as a control to verify that conventional operator-learning frameworks do not attain machine precision on these problems.
Furthermore, we note that the benchmark methods are trained in the low-data regime detailed in the following subsections, justifying the varying performance of the baselines. We also note that the vanilla FNO and vanilla TNO do not support the different input and output domains which arise in our benchmark inverse problems as detailed in \Cref{sec:inverse_problems_numerics}; hence, we do not apply these benchmarks to the inverse problems we test here. \Cref{tab:summary-perf} gives a performance comparison between these benchmark operator learning methods and our proposed \textsc{CHONKNORIS} approach.
The details of the benchmark problems are outlined in the remainder of
this section.
Additional details on the GP baselines are given in \Cref{sec:gp_benchmarks}.

\begin{table}[h!]
    \centering
    
    \begin{tabular}{l|cccc}
        & GP & FNO & TNO & \textsc{CHONKNORIS} (ours) \\ 
        \hline 
        Nonlinear elliptic & 5.1e-6 & 1.3e-3 & 5.7e-3 & 8.9e-16\\ 
        Burgers' & 1.1e-1 & 8.8e-3 & 2.2e-2 & 5.1e-16\\
        Nonlinear Darcy & 1.8e-3 & 4.8e-3 & 3.9e-3 & 9.6e-16\\
        Calder\`{o}n & 1.6e-2 & -- & -- & 3.2e-15\\ 
        Inverse wave scattering & 1.5e-2 & -- & -- & 9.2e-13\\
        Seismic imaging $5 \times 5$ & 2.3e{-2} & -- & -- & 2.0e-14\\
        Seismic imaging $7 \times 7$ &  4.4e{-2}& -- & -- &  3.0e-12\\
        Seismic imaging $10 \times 10$ &  5.7e{-2}& -- & -- & 1.2e-03\\
    \end{tabular}
    \caption{Summary of median relative $L^2$ losses over 
    multiple forward and inverse problem benchmarks.}
    \label{tab:summary-perf}
\end{table}

\subsection{Forward Problems} \label{sec:forward_problems}

\Cref{table:forward_pde_params} summarizes the forward problems we consider and their parameterizations. 
The following subsections provide additional details on each of these forward problems. 

\begin{table}[ht!]
    \centering
  
    \resizebox{\columnwidth}{!}{
    \begin{tabular}{r | c c c}
        forward problem & Nonlinear Elliptic 1D  & Burgers' 1D time-dependent & Darcy 2D \\
        \hline 
        $u$ distrib. & GP-periodic & sum of sines & GP-invLaplacian \\
        FD grid & $N_x = 63$ & $(N_t,N_x) = (151,127)$ & $(N_{x_1},N_{x_2}) = (20,20)$ \\
        $(R_\mathrm{train},R_\mathrm{val})$ & $(896,128)$ & $(448,64)$ & $(896,128)$ \\
        $(n_\mathrm{warm},\lambda_\mathrm{flow},\lambda_{\text{train}})$ & $(5,0,0)$ & $(5,0,10^{-2})$ & $(6,0,10^{-3})$ \\
        Hessian model & GP-Gaussian & MLP-Tanh $(127,500,1000,8128)$ & 
        GP-Gaussian \\
        Section & \Cref{sec:nonlinear_elliptic_pde_1d} & \Cref{sec:burgers_eq} & \Cref{sec:numerics:DarcyNonlinear} \\
        \hline 
        inverse problem & Calder\`{o}n  & Inverse Wave Scattering & Seismic Imaging FWI \\
        \hline 
        $u$ distrib. & GP-invLaplacian & GP-invLaplacian  & OpenFWI dataset \citep{deng2022openfwi} \\
        FD grid & $(N_{x_1},N_{x_2})=(9,9)$ & $(N_{x_1},N_{x_2})=(7,7)$ & $(N_{x_1},N_{x_2}) \in \{(5,5),(7,7),(10,10),(14,14)\}$ \\
        $(R_\mathrm{train},R_\mathrm{val})$ & $(7500,2500)$ & $(9750,250)$ & $(800,200)$ \\
        $(n_\mathrm{warm},\lambda_\mathrm{flow},\lambda_\mathrm{train})$ & $(0, 10^{-10},10^{-10})$ & $(0, 10^{-4}, 10^{-4})$ & $(400,\text{adaptive},\text{adaptive})$ \\
        Hessian model & GP--Gaussian & GP--Gaussian & GP--Gaussian \\
        Section & \Cref{sec:calderon} & \Cref{sec:wave_scattering} & \Cref{sec:seismic_imaging}
    \end{tabular}
    }
    \caption{Forward and inverse problem parameters. $u$ distrib. is the distribution of random coefficients. FD grid contains the regular grid sizes for the finite difference scheme. $(R_\mathrm{train},R_\mathrm{val})$ are the number of training and validation realizations respectively. $n_\mathrm{warm}$ denotes the number of Newton--Kantorovich iterations used to generate training data along the flow.  $\lambda_\mathrm{flow}$ is the relaxation used in the NK method, while $\lambda_{\text{train}}$ is the relaxation shown to the Hessian prediction model. \textsc{CHONKNORIS} training is performed with $\Lambda_\text{train} = \{\lambda_\text{flow}\}$ in the context of \Cref{alg:UnifiedNORIS}. GP-kernel denotes a Gaussian process with the given kernel, with the invLaplacian kernel given by $5(- \Delta + 1/100)^{-2}$ where $-\Delta$ denotes the Laplacian. MLP-nonlinearity is a multi-layer perceptron neural network architecture where the tuple specifies layer sizes, including input and output layers, and the given nonlinearity is applied to all hidden layers. Note that for Burgers' equation, $n_\text{warm} = 5$ NK steps were run per time step.}
    \label{table:forward_pde_params}
\end{table}

\subsubsection{\textbf{Nonlinear Elliptic Equation}}\label{sec:nonlinear_elliptic_pde_1d}

The first example is the 1D nonlinear elliptic PDE \eqref{eq:example_elliptic_pde} with $\kappa=50$ as introduced in \Cref{sec:operator_learning}. 

The differential operator is given in \eqref{eq:elliptic_pde_calF} and the corresponding solution operator is given by $\mathcal{G}(u)=v$ mapping the right hand side to the solution of the boundary value problem \Cref{sec:operator_learning}. 
The random coefficient $u$ is sampled from a zero-mean GP with a periodic kernel 
$$K(x,x') = \exp(-2/\ell \sin^2(\pi/p (x-x')))$$
with period length $p=1/2$ and lengthscale $\ell=10$. 
Here we fit two operator learning models. The first is an end-to-end operator learning model whose prediction is used as an initial guess for the NK/\textsc{CHONKNORIS} method. The second is our \textsc{CHONKNORIS} predictor model for the Cholesky factor. For both models we use vector-valued GP regression with Mat\'ern kernels with smoothness parameter $\nu = 5/2$ (which we abbreviate as $5/2$ Mat\'ern) 
and squared exponential kernels, all with tuned lengthscales.

\subsubsection{\textbf{Burgers' Equation}}
\label{sec:burgers_eq}

The second example is the 1D time-dependent Burgers' equation, supplemented with periodic boundary conditions and a random initial condition, which may be written as 
\begin{equation} \label{eq:Burgers'}
    \begin{cases}
        \partial_t f = \nu \partial_{xx} f - f \partial_x f, & (x,t) \in \mathbb{T} \times [0,T], \\
        f(0,t) = f(1,t), & t \in [0,T], \\
        f(x,0) = f_0(x), & x \in \mathbb{T}
    \end{cases}.
\end{equation}
Here $\nu = 1/50$, $f_0$ is the initial condition and $\partial_t$, $\partial_x$, $\partial_{xx}$ are partial derivatives. Here, we discretize the PDE in time and learn the solution operator of the time-discrete problem. More precisely, we discretize $[0,T]$ using the $M$-point uniform grid $\{t_i\}_{i=0}^{M-1} := \{i\Delta t\}_{i=0}^{M-1}$, $\Delta t = T/(M-1)$, and apply an implicit Euler time discretization leading to
$$\frac{f^{i+1}(x) - f^i(x)}{\Delta t} = \nu \partial_{xx} f^{i+1} (x) - f^{i+1}(x) \partial_{x} f^{i+1}(x)$$
where $f^{i}(x) = f(x,t_i)$ and $\Delta t = T/(M-1)$. Thus, the time marching discretization scheme defines the next time step $f^{i+1}$ given $f^{i}$ so that $\mathcal{F}(f^{i},f^{i+1})=0$ with 
\begin{equation} \label{eq:burgers_calF}
    \mathcal{F}(u,v)=v - \Delta t \left(\nu \partial_{xx} v - v \partial_{x} v\right) - u \qquad\text{and}\qquad
    \frac{\delta \mathcal{F}}{\delta v}(u,v)[h] = h - \Delta t\left(\nu \partial_{xx} h - \partial_{x} v h - v \partial_{x} h\right).
\end{equation}
Notice that $\delta \mathcal{F} / \delta v$ does \emph{not} depend on $u$. The resulting solution operator is given by $\mathcal{G}(u)=v$ mapping the previous time step $u=f^{i}$ to the next time step $v=f^{i+1}$. Here, \textsc{CHONKNORIS} approximates the solution operator $\mathcal{G}$ which in turn is used iteratively to obtain the solution for all time steps.
We take our random initial condition to be 
$$f_0(x) = \sum_{k=1}^3 a_k \sin(\pi k x), \qquad (a_1,\dots,a_3) \sim \mathcal{N}(0,1).$$

\subsubsection{\textbf{Nonlinear Darcy Flow}}\label{sec:numerics:DarcyNonlinear}

The third example is the 2D Darcy flow equation, supplemented with homogeneous Dirichlet boundary conditions, which may be written as 
\begin{equation} \label{eq:Darcy}
\begin{cases}
    - \nabla \cdot (e^u \nabla v) +  v^3 = f, & x \in [0,1]^2 \\
    v = 0, & x \in \partial [0,1]^2
\end{cases}
\end{equation}
with forcing term $f$ and conductivity $e^u$. Expanding $- \nabla \cdot (e^u \nabla v) = - e^u [\nabla u \cdot \nabla v + \Delta v]$, we have,  
\begin{equation} \label{eq:Darcy_calF}
    \mathcal{F}\left(u,v\right) = -e^u [\nabla u \cdot \nabla v + \Delta v]+\kappa v^3 - f \quad\text{and}\quad 
    \left[\frac{\delta \mathcal{F}}{\delta v}(u,v)\right](h) = -e^u[\nabla u \cdot \nabla h + \Delta h]+3 \kappa v^2 h.
\end{equation} Here, the solution operator is given by $\mathcal{G}(u,f)=v$ which maps the forcing $f$ and the conductivity $u$ to the solution $v$. In that sense, \textsc{CHONKNORIS} is learning a parametric family of solution operators $\mathcal{G}_u(\cdot)=\mathcal{G}(u,\cdot)$.
We take the fixed forcing term $f$ to be a single draw from a zero-mean GP with a $5/2$ Mat\'ern kernel and constant lengthscale of $3/10$ across both dimension.

\subsubsection{Summary of the numerical results for \textsc{CHONKNORIS} on the forward problems} \label{sec:CHONKNORIS_numerics}

We used \textsc{CHONKNORIS} to emulate and solve the three problems above. The domains were discretized using regular grids, and derivatives were approximated using standard finite differences. A traditional NK solver was used to generate the training data, and the reference solutions with which the errors were computed. The Cholesky factors for the nonlinear elliptic PDE in \Cref{sec:nonlinear_elliptic_pde_1d} and the Darcy flow problem in \Cref{sec:numerics:DarcyNonlinear} were predicted using a GP, while for the Burgers' equation in  \Cref{sec:burgers_eq} we used a Multi-Layer Perceptron (MLP). 

\begin{figure}[ht]
    \centering
    \includegraphics[width=1\linewidth]{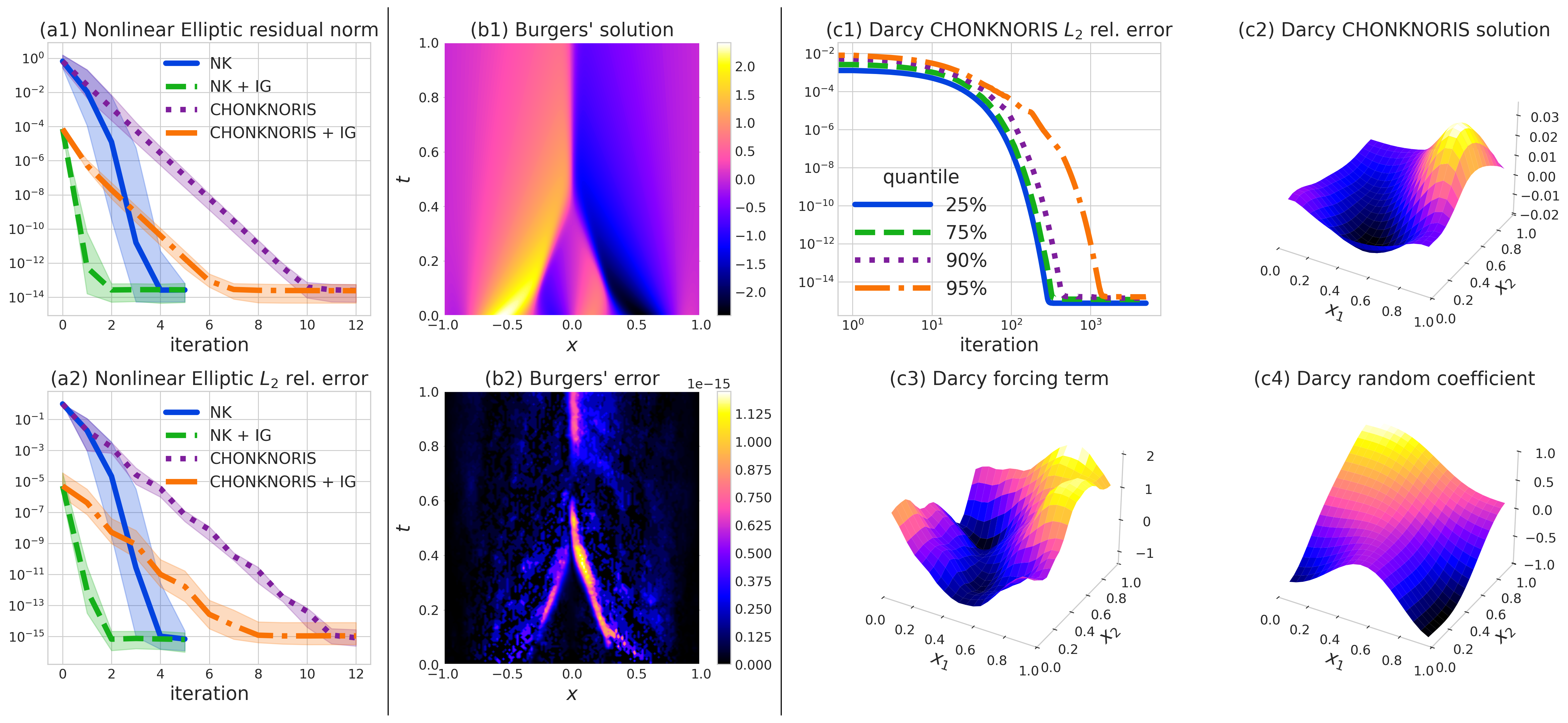}
    \caption{
    Forward problems. (a) Results for the nonlinear elliptic PDE problem. Quantiles of $10\%-90\%$ are shown across test realizations. Our \textsc{CHONKNORIS} method is able to achieve machine precision accuracy in around $10$ iterations. (b) Results for Burgers' equation. \textsc{CHONKNORIS} was able to achieve machine precision error in recovering the discretized solution which contained shocks.
    (c) Results for the Darcy flow PDE: (c1) shows that more challenging realizations require more \textsc{CHONKNORIS} iterations. (c4) shows a single realization of the random coefficient with the corresponding solution in (c2). (c3) shows the fixed forcing term.}
    \label{fig:forward_problems_complete}
\end{figure}

In all tested instances of the three problems, \textsc{CHONKNORIS} consistently achieves machine precision error. We study the speed of convergence in the nonlinear elliptic PDE in \Cref{sec:nonlinear_elliptic_pde_1d}, see \Cref{fig:forward_problems_complete} (a), and found that \textsc{CHONKNORIS} starting from an initial guess of $0$ typically converges in $10$ iterations while predicting the approximate Hessian, whereas the traditional NK solver needs only $4$ iterations using the precise approximate Hessian. We also observed that the convergence of the RMSE residual $\lVert \mathcal{F}(u,v_n) \rVert$, which we could track, follows the desired convergence of $L^2$ relative errors $\lVert v_n- v \rVert$ to machine precision (which is not generally known except when the true solution $v$ is known as is the case here). We observed that providing an initial guess from an end-to-end operator learning approach gives modest speedups of one to two iterations to both the NK and \textsc{CHONKNORIS} iterations.

We tested the robustness of our method with the Burgers' problem in \Cref{sec:burgers_eq}, see \Cref{fig:forward_problems_complete} (b), and found that our method can achieve machine precision even when solutions contain shocks. Finally, the Darcy flow problem in \Cref{sec:numerics:DarcyNonlinear} was the most challenging of the forward benchmarks as it required a lot of data to learn the parametric Cholesky factors. This experiment showcases the capacity of our method to utilize additional compute to achieve convergence even in difficult settings. As shown in \Cref{fig:forward_problems_complete} (c), within $10$ iterations \textsc{CHONKNORIS} usually reaches $L^2$ relative errors on the order of $10^{-3}$, putting it on par with existing operator learning models. After $100$ iterations, errors are typically on the order of $10^{-6}$ or better. Continuing to  increase the number of \textsc{CHONKNORIS} iterations to $1000$ enables convergence to machine precision in $95\%$ of all cases. In this example, 
we observed that realizations from a rougher distribution require significantly more \textsc{CHONKNORIS} iterations for exact recovery. 

\subsubsection{\textbf{Klein--Gordon and Sine--Gordon}}
\label{sec:sine_gordon}
For our last two  forward problem examples, we consider the 1D  Klein--Gordon and the Sine--Gordon equations. We will use these problems as a held out validation problem for 
testing the generalization error of \textsc{FONKNORIS}. When supplemented with initial and boundary condition, both PDEs take the form of 
\begin{equation} \label{eq:Sine-Klein--Gordon}
    \begin{cases} 
        \partial_{tt} f = \kappa_1 \partial_{xx} f - \kappa_2\tau(f), & (x,t) \in \mathbb{T} \times [0,T], \\
        f(0,t) = f(1,t), & t \in [0,T], \\
        f(x,0) = f_0(x), & x \in \mathbb{T}
    \end{cases}.
\end{equation}
Here $x \in \Omega = \mathbb{T}$ and $t \in [0,T]$, $f_0$ is the initial condition, and $\tau$ is a nonlinearity. For Klein--Gordon, $\tau(f)=f^3$, $\kappa_1=0.1$, $\kappa_2=10$, while for Sine--Gordon $\tau(f)=\sin(f)$ and $\kappa_1=\kappa_2=1$. Similar to the time marching scheme for Burgers' equation in \Cref{sec:burgers_eq}, we discretize $[0,T]$ and approximate $\partial_{tt}f \approx(f^{i+2}-2f^{i+1}+f^{i})/\Delta t^2$. Knowing the state $u:=(u_1,u_2):=(f^{i+1},f^i)$, we want to compute the next time step $v:=f^{i+2}$ defined as $\mathcal{F}((f^{i+1},f^i),f^{i+2})=0$ with
\begin{align}
    \mathcal{F}(u,v) &:= v - 2 u_1 + u_2 - (\Delta t)^2 (\nu \partial_{xx} v-\kappa_2\tau(v)), \label{eq:sk_gordon_calF}\\
    \left[\frac{\delta \mathcal{F}}{\delta v}(u,v)\right](h) &= h - (\Delta t)^2\left(\kappa_1 \partial_{xx}  h-\kappa_2 \frac{\delta \tau}{\delta v} (v)[h] \right), \label{eq:sk_gordon_calF_deriv}
\end{align} where $u=(u_1,u_2)$.
For Klein--Gordon, $\frac{\delta \tau}{\delta v}(v)[h]=3v^2 h$, and for Sine--Gordon $\frac{\delta \tau}{\delta v} (v)[h]=\cos(v)h$. We note again, that $\delta \mathcal{F}/\delta v$ is independent of $u$. The resulting solution operators are in both cases given by $\mathcal{G}(u)=v$ which map the previous time steps $u=(u_1,u_2)=(f^{i+1},f^i)$ to the next time step $v=f^{i+2}$. Again, \textsc{CHONKNORIS} approximates the solution operator $\mathcal{G}$ which is used in the very same  time marching scheme to obtain the solution for all time steps.

\begin{table}[H]
    \centering
    \begin{tabular}{|c|l|ccc|}
    \hline
        \textsc{FONKNORIS} & Partial differential equation & $a$ & $b$ & $c$ \\
        \hline 
        \multirow{3}{*}{training PDEs} & Nonlinear elliptic \eqref{eq:elliptic_pde_calF} & $-1$ & $0$ & $3 \kappa v^2$ \\
        & Burgers' \eqref{eq:burgers_calF} & $-(\Delta t) \nu$ & $(\Delta t)v$ & $1 + (\Delta t) \nabla v$ \\
        & Nonlinear Darcy flow \eqref{eq:Darcy_calF} & $-e^u$ & $-e^u \nabla u$ & $3 \kappa v^2$\\
        \hdashline
        \multirow{ 2}{*}{testing PDEs} & Sine--Gordon \eqref{eq:sk_gordon_calF}/\eqref{eq:sk_gordon_calF_deriv} & $-\kappa_1 (\Delta t)^2$ & $0$ & $1+ \kappa_2 (\Delta t)^2 \cos(v)$\\
        & Klein--Gordon \eqref{eq:sk_gordon_calF}/\eqref{eq:sk_gordon_calF_deriv} & $-\kappa_1 (\Delta t)^2$ & $0$ & $1 + 3 \kappa_2 (\Delta t)^2 v^2$ \\
        \hline
    \end{tabular}
    \caption{\textsc{FONKNORIS} coefficients.}
    \label{tab:fonknoris_abc}
\end{table}

\begin{figure}[!ht]
    \centering
    \includegraphics[width=1\linewidth]{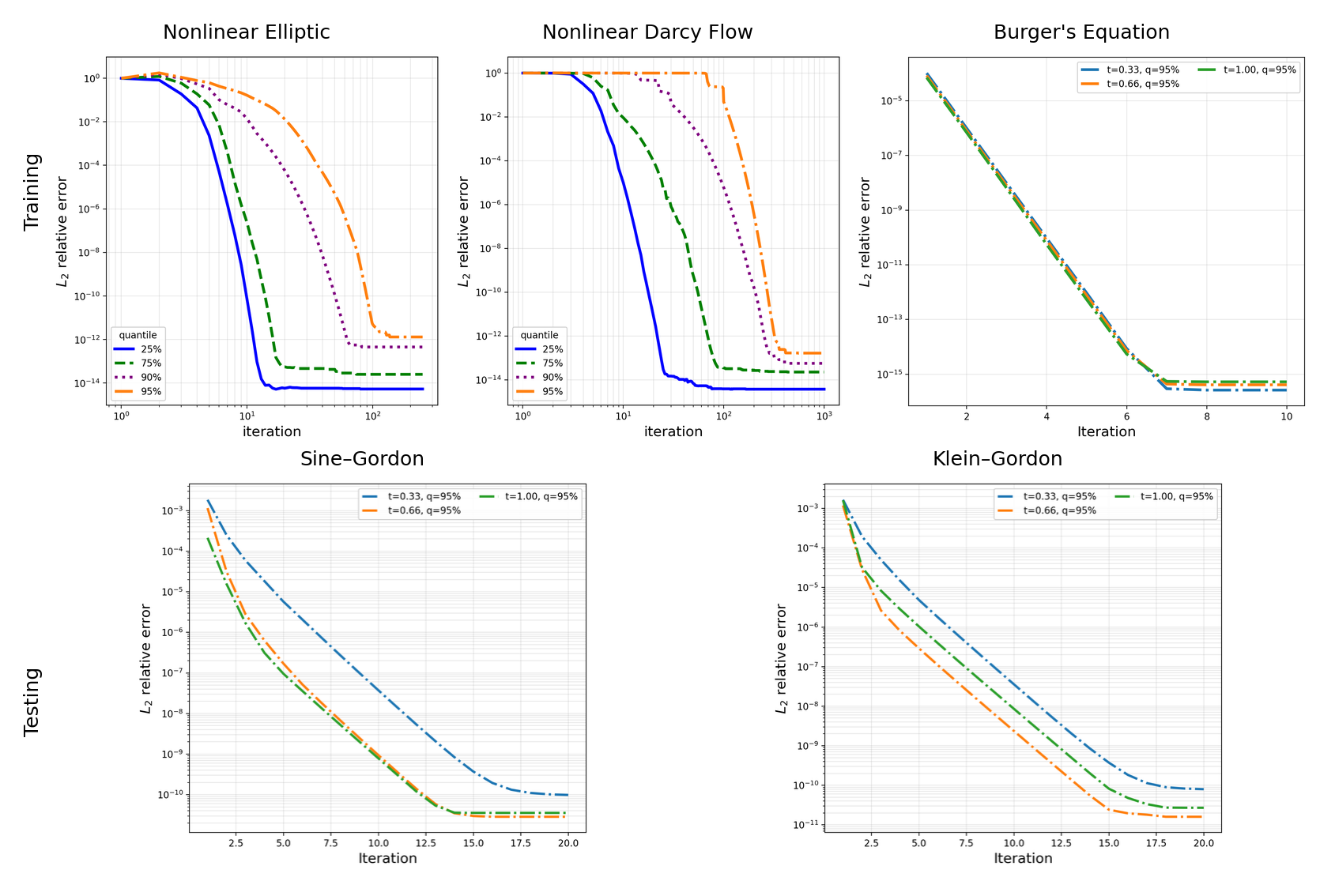}
    \caption{
    Quantiles of predictions of \textsc{FONKNORIS} for 100 realizations of initial conditions, external forces, and conductivities using a mixture of experts consisting of GPs for nonlinear elliptic, nonlinear darcy flow, and Burgers' equation and testing it for withheld Sine--Gordon and Klein--Gordon equations.}
    \label{fig:fonknoris}
\end{figure}

\subsection{Foundation Model - \textsc{FONKNORIS}} \label{sec:foundation_modeling_numerics}

For the \textsc{FONKNORIS} model, we train sub-model experts, each specific to a given PDE, and aggregate  
individual \textsc{CHONKNORIS} sub-models within mixture of experts \cite{Rulliere2018}. The aggregation of the sub-models is done by the so-called nested Kriging method \cite{Rulliere2018}, see also
\Cref{sec:nested_kriging}. This method relies on the fact that each sub-model has the same underlying Gaussian process and thus share the same kernel. In our \textsc{FONKNORIS} prediction framework, we employ Gaussian processes to model and predict the behavior of specific PDEs for each expert model. Constructing a single foundational Gaussian process model capable of representing a large class of PDEs would require an extensive dataset, which would quickly exceed typical computational and memory limits. To overcome this limitation, we train each expert model independently and subsequently aggregate the individual experts within a foundational (meta) model. 

\subsubsection{Data Generation and Training}
In our numerical experiments for \textsc{FONKNORIS}, we trained separate instances of the model $\hat{\mathcal{R}}_\theta$ for the one-dimensional nonlinear elliptic equation, Burgers' equation, and the one-dimensional nonlinear Darcy flow, using Mat\'{e}rn kernels with learned length scales $\ell_1, \ell_2,$ and $\ell_3$ for the input distributions $\mu$. For the aggregated model, we used the arithmetic mean of the length scales, $\ell_{\mathcal{A}}=\frac{1}{p}\sum_{i=1}^p \ell_i.$
The Sine--Gordon and Klein--Gordon equations were held out for testing. All PDEs were discretized with $N_x=64$ collocation points. In the Newton--Kantorovich warm-up step of Algorithm~\ref{alg:UnifiedNORIS} we used a fixed flow relaxation parameter $\lambda_{\mathrm{flow}} = 0.01$ and the same value in the training relaxation set, $\lambda_{\text{train}} = 0.01.$

For the nonlinear elliptic equation and Burgers' equation, we generated $M_{\mathrm{ell}} = 5000$ and $M_{\mathrm{Burg}} = 5000$ offline realizations, respectively. The realizations for the nonlinear elliptic equation were generated by drawing $1000$ external forcings $f$ from a periodic kernel defining $\mu$ and running $n_{\text{warm}} = 5$ Newton--Kantorovich iterations. The realizations for Burgers' equation were generated by drawing $1000$ initial conditions $u_0$ from a periodic kernel and again running $n_{\text{warm}} = 5$ Newton--Kantorovich iterations. For the nonlinear Darcy flow, we generated $M_{\mathrm{Darcy}} = 10000$ realizations by drawing $2000$ pairs of conductivities $a$ and external forcings $f$ from a periodic kernel and running $n_{\text{warm}} = 5$ Newton--Kantorovich iterations for each pair $(a_i,f_i)$. The aggregation model was tested on the Klein--Gordon and Sine--Gordon equations by drawing realizations of the initial condition $u_0$ from a periodic kernel and choosing $v_0 = 0$ in the online phase of Algorithm \ref{alg:UnifiedNORIS}. The data for training and testing were generated from the same distributions.

\subsubsection{Summary of the numerical results for \textsc{FONKNORIS} on the forward problems} \label{sec:FONNORIS_numerics}

We applied \textsc{FONKNORIS} to the above problems, using the nonlinear elliptic, the Burgers', and the nonlinear Darcy flow equations as training problems, and testing on all five problems, including the Klein--Gordon and Sine--Gordon equations. In \Cref{fig:fonknoris}, we see that the aggregated model $M(x)$ not only achieves machine precision on the trained instances, but also achieves near machine precision for the withheld Klein--Gordon and Sine--Gordon equation. \Cref{tab:fonknoris_abc} shows the \textsc{FONKNORIS} coefficients for the aforementioned forward problems. The advantage of this approach is its simplicity and generating training data does not rely on a specific PDE as the coefficients $a,b,c$ can be generated from a certain distribution, optimally from various distributions. The main difficulty is that the model can become very large and the coefficients $a,b,c$ can easily get out of distribution when computing each Newton--Kantorovich step; one deals with the curse of dimensionality as the generated data might not inherit the statistics of a given PDE. Therefore, we generate the training data from the first three problems and generalize it to all problems including the Sine--Gordon and Klein--Gordon equation. 
Here, our mixture-of-experts Gaussian process model was capable of handling the large scale data necessary to generalize across PDE problems. 
This ability to generalize to unseen problems at machine precision is a state-of-the-art achievement that is due to the unique design of the \textsc{FONKNORIS} model.

\subsection{Inverse Problems} \label{sec:inverse_problems_numerics}

In this section, we introduce various inverse problems which will be used to further test the performance of  \textsc{CHONKNORIS} as an emulator for PDE constrained optimization. These include the Calder\`{o}n problem (\Cref{sec:calderon}), the inverse wave scattering problem (\Cref{sec:wave_scattering}), and a problem in seismic imaging full waveform inversion (\Cref{sec:seismic_imaging}).  Specific parameterizations for each of these problems are given in \Cref{table:forward_pde_params}. The following subsections further detail our setup.

\subsubsection{\textbf{The Calder\`{o}n problem}}\label{sec:calderon}

Let $\Omega =[0,1]^2$ and assume $v \in L^{\infty}(\Omega)$ is a real-valued conductivity with $v(x) > 0$ for almost every  $x \in \Omega$. 
Consider the boundary value problem 
\begin{equation} \label{eq:Calderon.PDE}
    \begin{cases}
        \nabla \cdot (v(x)\nabla c(x)) = 0 & \text{ for }x\in \Omega,\\
        c(x)=g(x) & \text{ for }x\in\partial \Omega, 
    \end{cases}
\end{equation} where on the boundary, we prescribe a boundary voltage pattern $g\in H^{\frac{1}{2}}(\partial \Omega)$. We define the \textit{Dirichlet-to-Neumann} (DtN) map $\Lambda_g: H^{\frac{1}{2}}(\partial \Omega)\rightarrow  H^{-\frac{1}{2}}(\partial \Omega)$ that maps the boundary voltage pattern $g$ to the current flux $
v\frac{\partial c}{\partial n}\bigg|_{\partial \Omega}$. Note that $\Lambda_g\in \mathcal{L}(H^{\frac{1}{2}}(\partial \Omega), H^{-\frac{1}{2}}(\partial \Omega))$ is a linear bounded operator. 
The \textit{Calder\`{o}n problem} is the task of recovering the conductivity $v$ from a given DtN map $\Lambda_g$. Denote with $\Tilde{\mathcal{F}}:L^\infty(\Omega)\rightarrow \mathcal{L}(H^{\frac{1}{2}}(\partial \Omega), H^{-\frac{1}{2}}(\partial \Omega))$ the forward operator that maps the conductivity to the Dirichlet-to-Neumann map. Then, the operator of interest is given by
\begin{align*}
    \mathcal{F}(\Lambda_g,v)= \Tilde{\mathcal{F}}(v)- \Lambda_g.
\end{align*} \textsc{CHONKNORIS} aims to approximate the operator $\mathcal{G}: \mathcal{L}(H^{\frac{1}{2}}(\partial \Omega), H^{-\frac{1}{2}}(\partial \Omega))\rightarrow L^\infty(\Omega)$ that maps the DtN map to the conductivity, i.e., $\mathcal{G}(\Lambda_g)=\Tilde{\mathcal{F}}^{-1}(\Lambda_g)=v$. We note that, in practice, we only have access to pairs of observations $(g_i, v \frac{\partial c_i}{\partial n}|_{\partial \Omega})_{i=1}^N$ on a finite number of sensor points on the boundary $\partial \Omega$, where $c_i$ denotes the solution to the \eqref{eq:Calderon.PDE} with conductivity $v$ and boundary condition prescribed by $g_i$. Thus the equation used in practice is \begin{equation*}
    \hat{\mathcal{F}}(u,v)=\left(\Tilde{\mathcal{F}}(v)g_i- v \frac{\partial c_i}{\partial n}|_{\partial \Omega}\right)_{i=1}^N=0,
\end{equation*} where $u=(g_i, v \frac{\partial c_i}{\partial n}|_{\partial \Omega})_{i=1}^N$.

We take the fixed forcing term to be a single draw from a zero-mean GP with $5/2$ Mat\'ern kernel and constant lengthscale of $3/10$ across both dimension. 

\subsubsection{\textbf{Inverse Wave Scattering}}\label{sec:wave_scattering}

Let $\Omega =[0,1]^2$ and assume that $a \in L^{\infty}(\Omega)$ is a real-valued material property  satisfying $a(x) > 0$ for almost every $x \in \Omega$. 
We study the following elliptic boundary value problem:
\begin{equation} \label{eq:wave.scattering.PDE}
    \begin{cases}
        -\Delta u(x) - \omega^2 a(x)u(x) = 0, & x \in \Omega, \\
        u(x) = g(x), & x \in \partial \Omega,
    \end{cases}
\end{equation}
where $g \in H^{\frac{1}{2}}(\partial \Omega)$ denotes the prescribed boundary excitation.

We define the \textit{Dirichlet-to-Neumann} (DtN) map 
$\Lambda_a: H^{\frac{1}{2}}(\partial \Omega)\rightarrow  H^{-\frac{1}{2}}(\partial \Omega)$
that maps the boundary input $g$ to the corresponding flux 
\[
\Lambda_a(g) = \frac{\partial u}{\partial n}\bigg|_{\partial \Omega}.
\]
Note that $\Lambda_a \in \mathcal{L}(H^{\frac{1}{2}}(\partial \Omega), H^{-\frac{1}{2}}(\partial \Omega))$ is a bounded linear operator depending on the material coefficient $a$.

The \textit{inverse wave scattering problem} is the task of recovering the coefficient $a$ from a given DtN map $\Lambda_a$.  
Denote with $\Tilde{\mathcal{F}}:L^\infty(\Omega)\rightarrow \mathcal{L}(H^{\frac{1}{2}}(\partial \Omega), H^{-\frac{1}{2}}(\partial \Omega))$ the forward operator that maps the material property $a$ to the DtN map.  
Then, the operator of interest is defined as
\begin{align*}
    \mathcal{F}(\Lambda_a, a) = \Tilde{\mathcal{F}}(a) - \Lambda_a.
\end{align*}
Again, \textsc{CHONKNORIS} aims to approximate the operator $\mathcal{G}: \mathcal{L}(H^{\frac{1}{2}}(\partial \Omega), H^{-\frac{1}{2}}(\partial \Omega)) \rightarrow L^\infty(\Omega)$ that maps the DtN map to the material property, i.e., $\mathcal{G}(\Lambda_a) = \Tilde{\mathcal{F}}^{-1}(\Lambda_a) = a$.

In practice, we only have access to finitely many boundary input-output pairs $(g_i, \frac{\partial u_i}{\partial n}|_{\partial \Omega})_{i=1}^N$ measured on sensor locations along $\partial \Omega$, where $u_i$ denotes the solution to \eqref{eq:wave.scattering.PDE} corresponding to the boundary excitation $g_i$. Hence, the equation used in practice is
\begin{equation*}
    \hat{\mathcal{F}}(u,a)=\left(\Tilde{\mathcal{F}}(a)g_i - \frac{\partial u_i}{\partial n}\bigg|_{\partial \Omega}\right)_{i=1}^N = 0,
\end{equation*}
where $u=(g_i, \frac{\partial u_i}{\partial n}|_{\partial \Omega})_{i=1}^N$.

We take the fixed forcing term to be a single draw from a zero-mean Gaussian process with $5/2$ Mat\'ern kernel and constant lengthscale of $2/10$ across both dimension.

\subsubsection{\textbf{Seismic Imaging}}\label{sec:seismic_imaging}

Let $\Omega = [0,1]^2$ and let $T > 0$ denote the final observation time. Assume that $v \in L^{\infty}(\Omega)$ is a real-valued velocity coefficient satisfying $v(x) > 0$ for almost every $x \in \Omega$.  
We study the following time-dependent acoustic wave equation:
\begin{equation} \label{eq:seismic.imaging.PDE}
    \begin{cases}
        \Delta p(t,x) - \frac{1}{v^2(x)} p_{tt}(t,x) = s(t,x), & (t,x) \in (0,T)\times\Omega, \\
        p(0,x) = 0, \quad p_t(0,x) = 0, & x \in \Omega,
    \end{cases}
\end{equation}
where $p$ denotes the pressure variation and $s$ represents the source term which we take to be a 
Ricker wavelet \cite{wang2015frequencies}.

The reflected and refracted wavefields are measured on the surface $\mathcal{S}:=\lbrace (x_1,x_2)\in \Omega\mid x_2=0\rbrace \subset \partial \Omega$, yielding the boundary observations $p|_{[0,T]\times\mathcal{S}}$. We define the \textit{Source-to-Receiver} (StR) map 
$\Lambda_v: L^2((0,T)\times\Omega) \rightarrow L^2(0,T; H^{\frac{1}{2}}(\mathcal{S}))$
that maps a source $s$ to the measured surface signal $\Lambda_v(s) = p|_{[0,T]\times\mathcal{S}}$. Note that $\Lambda_v \in \mathcal{L}(L^2((0,T)\times\Omega), L^2(0,T; H^{\frac{1}{2}}(\mathcal{S}))$ is a bounded linear operator depending on the velocity coefficient $v$.

The \textit{seismic inverse problem} or \textit{full wave inversion problem} is the task of recovering the coefficient $v$ from a given Source-to-Receiver map $\Lambda_v$ \cite{virieux2009overview}.
Denote with $\Tilde{\mathcal{F}}: L^\infty(\Omega)\rightarrow \mathcal{L}(L^2((0,T)\times\Omega), L^2(0,T; H^{\frac{1}{2}}(\mathcal{S})))$ the forward operator that maps the velocity coefficient $v$ to the corresponding StR map.  
Then, the operator of interest is defined as
\begin{align*}
    \mathcal{F}(\Lambda_v, v) = \Tilde{\mathcal{F}}(v) - \Lambda_v(s).
\end{align*}
Similar to the inverse wave scattering and Calder\'on problems, \textsc{CHONKNORIS} aims to approximate the operator 
\[
\mathcal{G}: \mathcal{L}(L^2((0,T)\times\Omega), L^2(0,T; H^{\frac{1}{2}}(\mathcal{S}))) \rightarrow L^\infty(\Omega),
\]
that maps the Source-to-Receiver map to the velocity coefficient, i.e., $\mathcal{G}(\Lambda_v) = \Tilde{\mathcal{F}}^{-1}(\Lambda_v) = v$.

\begin{figure}[htp]
    \centering
    \includegraphics[width=1\linewidth]{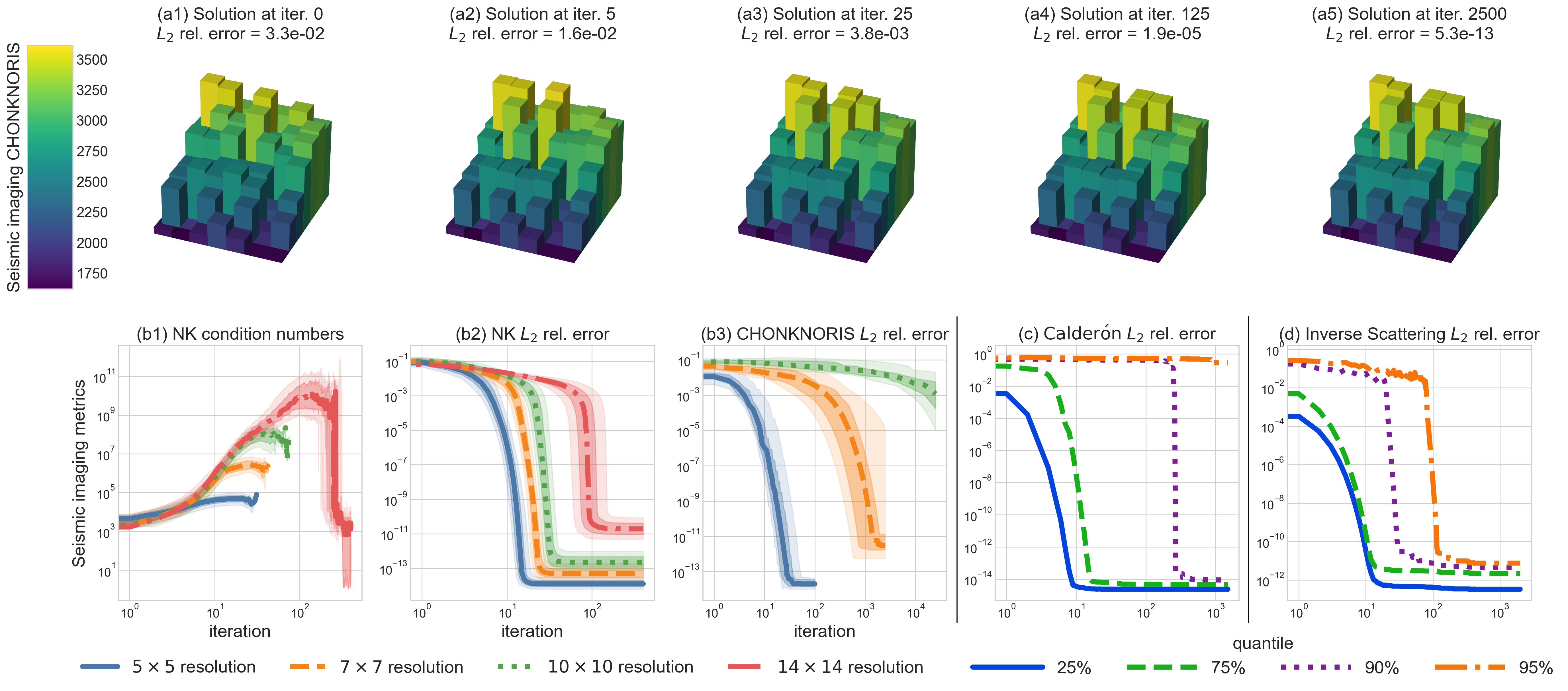}
    \caption{Inverse Problems. 
(a) Results for the seismic imaging problem, showing the iterative solutions and the relative $L^2$ error across \textsc{CHONKNORIS} iteration. 
(b) For the seismic imaging problem, evolution of the adaptive regularization term in the Newton--Kantorovich iterations for different resolutions, and comparison of the relative $L^2$ error between the Newton--Kantorovich method and our \textsc{CHONKNORIS} method.
(c) Results for the Calder\'on problem. 
(d) Results for the inverse wave scattering problem.
}
\label{fig:inverse_problems_complete}
\end{figure}

In practice, we only have access to finitely many input-output pairs $(s_i, p_i|_{[0,T]\times\mathcal{S}})_{i=1}^N$ measured at discrete sensor locations along $\mathcal{S}$, where $p_i$ denotes the solution to \eqref{eq:seismic.imaging.PDE} corresponding to the source $s_i$. Hence, the equation used in practice is
\begin{equation*}
    \hat{\mathcal{F}}(u,v)=\left(\Tilde{\mathcal{F}}(v)s_i - p_i|_{[0,T]\times\mathcal{S}}\right)_{i=1}^N = 0,
\end{equation*}
where $u = (s_i, p_i|_{[0,T]\times\mathcal{S}})_{i=1}^N$.
%
We use data from the OpenFWI dataset \citep{deng2022openfwi} and the forward solver from \cite{wang2023_2d28_fd_acoustic_modeling_lab} (a $2-4$ finite difference scheme with $2$nd-order accuracy in time and $4$th-order in space). 
Convergence of the exact NK method for a single $14 \times 14$ resolution velocity map is shown in \Cref{fig:inverse_problems_complete} (a1-a5).

While our previous experiments predicted the inverse Cholesky of the approximate Hessian, for this problem we obtained better performance by  directly predicting the Cholesky factors of the approximate Hessian and then use triangular solves to determine each increment. We note this does not change the computational complexity of our method. We also found it necessary to simultaneously tune both the learning rate $\alpha$ in the line search and the Tikhonov relaxation $\lambda$ in order to converge to machine precision in a reasonable number of iterations, see \Cref{alg:UnifiedNORIS}. The tuning scheme we used is described in \Cref{se:Learning_rate_relaxation}.

\subsubsection{Summary of the numerical results for \textsc{CHONKNORIS} on the Inverse Problems}

We use \textsc{CHONKNORIS} to solve the three inverse problems above. The domains are discretized using regular grids. A traditional NK solver is used to generate training data where the converged iterates are used as reference solutions. 
These inverse problems are significantly harder to solve than the forward problems due to the non-locality of the operator $\mathcal{F}$ and the ill-posedness of the  inverse problem. The results for the Calder\'on problem in  \Cref{fig:inverse_problems_complete} (d) show that running the \textsc{CHONKNORIS} method for $10^3$ iterations was able to recover solutions to machine precision for over $75\%$ of withheld test realizations.  The results for the inverse wave scattering problem in \Cref{fig:inverse_problems_complete} (e) shows that running the \textsc{CHONKNORIS} method for $40$ iterations is able to recover solutions to machine precision. 
For the rough velocity maps encountered in the seismic imaging problem, increasing the resolution increases the condition numbers of the relaxed approximate Hessian. For example, \Cref{fig:inverse_problems_complete} shows that the seismic imaging problem with just a $10 \times 10$ velocity map resolution encounters condition numbers around $10^{8}$ for near-convergence iterations. \textsc{CHONKNORIS} is unable to exactly predict the Cholesky factor of ill-conditioned matrices, and thus resorts to inferring gradient descent steps which can make \textsc{CHONKNORIS} inference slow to converge. As expected, we also observe that the relaxation is decreased as NK nears convergence, indicating a smooth transition from gradient descent to Gauss--Newton updates. This behavior is more difficult to replicate with \textsc{CHONKNORIS} as the approximate Hessian near the solution is ill-conditioned and therefore difficult for \textsc{CHONKNORIS} to predict.

\section{Theoretical results} \label{sec:theoretical_guarantees}
In this section, we want to present several results on the 
convergence properties of the \textsc{CHONKNORIS} and \textsc{FONKNORIS} algorithms. This is accomplished by combining an inexact Newton--Kantorovich method with Tikhonov-regularized inverse and the kernel-based operator learning method.

\subsection{Analytic Setting}\label{subsec:analytic_setting_varlam}
Assume that we are in the operator learning framework of \Cref{secoplepb}, i.e., let $(\mathcal U, \langle\cdot,\cdot\rangle_{\mathcal U},\Vert \cdot \Vert_\U)$, $(\mathcal V, \langle\cdot,\cdot\rangle_{\mathcal V},\Vert \cdot \Vert_\V)$, and $(\mathcal W ,\langle\cdot,\cdot\rangle_{\mathcal W},\Vert \cdot \Vert_\W)$ be separable Hilbert spaces. For notational convenience, we suppress the dependence of $\F$ on the first argument, i.e., for each fixed $u\in\mathcal U$, we define $\F(\cdot):=\F(u,\cdot):\mathcal V\to\mathcal W$ and its Fr\'echet derivative
$\F'(v):=\frac{\delta \F}{\delta v}(u,v)\in\mathcal L(\mathcal V,\mathcal W)$ with Hilbert adjoint $\F'(v)^{*}\in\mathcal L(\mathcal W,\mathcal V)$.

For $\lambda>0$ and $v\in\mathcal V$, define the \emph{Tikhonov resolvent} and the \emph{Tikhonov right inverse}
\[
R_\lambda(v):=\big(\lambda I_{\mathcal V}+\F'(v)^{*}\F'(v)\big)^{-1}\in\mathcal L(\mathcal V,\mathcal V),
\qquad
B_\lambda(v):=R_\lambda(v)\,\F'(v)^{*}\in\mathcal L(\mathcal W,\mathcal V).
\]
Let $\widehat R_\lambda(v)\in\mathcal L(\mathcal V,\mathcal V)$ be a (learned) surrogate and set
$\widehat B_\lambda(v):=\widehat R_\lambda(v)\,\F'(v)^{*}$.
Given $v_0\in\mathcal V$, consider the iteration
\[
v_{k+1}=v_k+\delta v_k,\qquad
\delta v_k:=-\,\widehat B_{\lambda_k}(v_k)\,\F(v_k)
     =-\,\widehat R_{\lambda_k}(v_k)\,\F'(v_k)^{*}\,\F(v_k),
\]
for a sequence of parameters $\lambda_k>0$. Furthermore, we make the following assumptions.

\begin{Assumption}\label{ass:NK}
Let $D\subset\mathcal V$ be open and convex and $\F:D\to\mathcal W$ be $C^1$.
Fix $v_0\in D$ and $R>0$ with $B:=\overline{B(v_0,R)}\subset D$.
Assume:
\begin{align*}
\text{\rm(A1)}\;& \F'(v_0)\ \text{is invertible and}\ \ \|\F'(v_0)^{-1}\|_{\mathcal L(\mathcal W,\mathcal V)}\le \beta\in \mathbb{R}_{>0},\,\\
\text{\rm(A2)}\;& \text{There exists } L>0 \text{ s.t. } \|\F'(u)-\F'(v)\|_{\mathcal L(\mathcal V,\mathcal W)}\le L\,\|u-v\|_\V \quad \forall\,u,v\in B,\\
\text{\rm(A3)}\;& M\ :=\ \sup_{v\in B}\ \|\F'(v)\|_{\mathcal L(\mathcal V,\mathcal W)}<+\infty,\quad \sigma_* := \inf_{v\in B}\ \sigma_{\min}\!\big(\F'(v)\big) = \inf_{v\in B}\ \|\F'(v)^{-1}\|_{\mathcal L(\mathcal W,\mathcal V)}^{-1}>0.
\end{align*}
\end{Assumption}

Now, we are in a position to state the main theorem on the Tikhonov-regularized inexact Newton--Kantorovich iteration method based on a classical inexact Newton--Kantorovich method \cite{steighaug1982inexact}. The proofs in this section are postponed to the appendix \Cref{sec:appendix_proofs}.

\begin{Theorem}[Tikhonov--inexact Newton--Kantorovich]\label{thm:main}
Let Assumption \ref{ass:NK} be satisfied.
Then, for each $v_k$ with $v_k\in\overline{B(v_0,R)}$, the linearized residual satisfies
\begin{equation}\label{eq:forcing_varlam}
\frac{\ \|\F'(v_k)\delta v_k+\F(v_k)\|_\W \ }{\ \|\F(v_k)\|_\W \ }
\ \le\
\frac{\lambda_k}{\lambda_k+\sigma_*^2} \ +\
M^2\,\varepsilon_{\lambda_k}
\ =:\ \theta_k,
\end{equation} where \[
\varepsilon_\lambda\ :=\ \sup_{v\in \overline{B(v_0,R)}}\ \big\|\,\widehat R_\lambda(v)-R_\lambda(v)\,\big\|_{\mathcal L(\mathcal V,\mathcal V)}
\] denotes the design/learning error.
If $\sup_k \theta_k\le \bar\theta<1$ and all iterates remain in $\overline{B(v_0,R)}$, then the inexact Newton--Kantorovich conclusions hold on $\overline{B(v_0,R)}$ with
\[
\widetilde L\ :=\ \frac{L}{1-\bar\theta},\qquad
\widetilde h\ :=\ \beta\,\widetilde L\,\eta,\qquad \eta:=\|\F'(v_0)^{-1}\F(v_0)\|_\V.
\]
In particular, if $\widetilde h\le \tfrac12$ and $t_*=(1-\sqrt{1-2\widetilde h})/(\beta\widetilde L)\le R$, then the iterates are well-defined, remain in $\overline{B(v_0,t_*)}$, and converge to the unique zero $v_*\in \overline{B(v_0,t_*)}$, with the Kantorovich majorant bounds
\[
\|v_k-v_*\|_\V \le t_*-t_k,\qquad
\|v_{k+1}-v_k\|_\V\le t_{k+1}-t_k,
\]
where $t_{k+1}=t_k-\dfrac{\phi(t_k)}{\phi'(t_k)}$ and $\phi(t)=\eta-t+\tfrac12\,\beta \widetilde L\,t^2$.

Moreover, if $\lambda_k\to 0$ and $\varepsilon_{\lambda_k}\to 0$, then $\theta_k\to 0$ and the local rate approaches the quadratic rate of exact Newton. 
\end{Theorem}

The following corollary makes the final statement of the preceding \Cref{thm:main} on the convergence more precise. In particular, it shows for which Tikhonov parameters and design errors we obtain linear, superlinear, quadratic convergence. 
\begin{Corollary}[Convergence Rates]\label{cor:rates1}
Let $e_k:=\|v_k-v_*\|_\V$. Under the hypotheses of \Cref{thm:main} and for all $k$ with $v_k\in\overline{B(v_0,R)}$,
\begin{equation}\label{eq:one-step1}
e_{k+1}\ \le\ \frac{1}{1-\bar\theta}\left(\frac{\beta L}{2}\,e_k^2\ +\ \bar\theta\,e_k\right).
\end{equation}
Then, there exist constants $C_1,C_2,C_3,C_4$ such that
\begin{enumerate}
\item[\textup{(i)}] If $\bar\theta\in(0,1)$ is fixed, then $\displaystyle \limsup_{k\to\infty}\frac{e_{k+1}}{e_k}\ \le\ \frac{\bar\theta}{1-\bar\theta}$ (at least $Q$-linear).
\item[\textup{(ii)}] If $\theta_k\to 0$, then $e_{k+1}\le \tfrac{\beta L}{2}\,e_k^2(1+o(1))$ and the rate is $Q$-superlinear.
\item[\textup{(iii)}] If $\theta_k\le C_1\,\|\F(v_k)\|_\W ^\alpha$ for some $\alpha>0$, then $e_{k+1}\le C_2\,e_k^{1+\alpha}+C_3e_k^2$ for $k$ large, so the local order is $\min\lbrace 2,1+\alpha\rbrace$.
\item[\textup{(iv)}] If $\theta_k=\mathcal O(\|\F(v_k)\|_\W)$, then $e_{k+1}\le C_4\,e_k^2$ and the convergence is $Q$-quadratic.
\end{enumerate}
\end{Corollary}

Based on the preceding corollary, we can propose a schedule for the Tikhonov parameter $\lambda_k$ and the design error $\varepsilon_{\lambda_k} $in order to achieve the desired convergence. 

\begin{Remark}[Scheduling $\lambda_k$ to reach superlinear/quadratic convergence]\label{prop:schedule}
Let $\theta_{\rm tik,k}=\dfrac{\lambda_k}{\lambda_k+\sigma_*^2}$ and $\theta_{\rm des,k}=M^2\,\varepsilon_{\lambda_k}$ so that $\theta_k=\theta_{\rm tik,k}+\theta_{\rm des,k}$.
\begin{enumerate}
\item[\textup{(a)}] If $\lambda_k\to 0$ and $\varepsilon_{\lambda_k}\to 0$, then $\theta_k\to 0$ and the rate is superlinear.
\item[\textup{(b)}] If there exist $c_1,c_2>0$ such that
\[
\lambda_k\ \le\ c_1\,\|\F(v_k)\|_\W,\qquad
\varepsilon_{\lambda_k}\ \le\ c_2\,\|\F(v_k)\|_\W,
\]
then $\theta_k=\mathcal O(\|\F(v_k)\|_\W)$ and the convergence is $Q$-quadratic.
\item[\textup{(c)}] Suppose for some $C>0$ one has the regularity proxy $\varepsilon_\lambda \approx C\,\lambda^{-2}$.
Consider $\phi(\lambda):=\dfrac{\lambda}{\lambda+\sigma_*^2}+C\,\lambda^{-2}$.
Then any $\lambda_k$ chosen near the minimizer $\lambda_*\ \asymp\ (C\,\sigma_*^2)^{1/3}$ balances Tikhonov bias and model error early on; subsequently decreasing $\lambda_k\downarrow 0$ (as the model improves and the iterates approach $v_*$) drives $\bar\theta\to 0$ and approaches quadratic convergence.
\end{enumerate}
\end{Remark}

\begin{Remark}[Variant: learning the full $B_\lambda(v)$ directly]
If the surrogate model directly learns $B_\lambda(v)$ and achieves
$\sup_{v\in B}\|\,\widehat B_\lambda(v)-B_\lambda(v)\|_{\mathcal L(\mathcal W,\mathcal V)}\le \varepsilon_\lambda$,
then the residual bound becomes
\[
\bar\theta\ \le\ \underbrace{\frac{\lambda}{\lambda+\sigma_*^2}}_{\theta_{\rm tik}}
\ +\ \underbrace{M\,\varepsilon_\lambda}_{\theta_{\rm des}},
\]
since $\|I-\F' \widehat B_\lambda\|\le \|I-\F'B_\lambda\|+\|\F'\|\,\|\widehat B_\lambda-B_\lambda\|$.
All conclusions of \Cref{thm:main} remain unchanged with this $\bar\theta$.
\end{Remark}

\begin{Example}[Application to nonlinear elliptic PDE]
\end{Example}
In this example, we want to apply the convergence theorem to the nonlinear elliptic PDE considered in Section \Cref{sec:nonlinear_elliptic_pde_1d}.
Let $\T=\mathbb{R}/\mathbb{Z}$ and
\[
\V:=H^1_{\mathrm{per},0}(\T)=\Bigl\{u\in H^1_{\mathrm{per}}(\T):\int_0^1 u=0\Bigr\},\qquad
\W:=\V^*\simeq H^{-1}_{\mathrm{per},0}(\T),
\]
with $\|u\|_{\V}:=\|u'\|_{L^2(0,1)}$.
Let $f\in L^2(0,1)$ be fixed and define
\[
\F:\V\to \W,\qquad \F(u)=-u''+u^3-f,\qquad
\F'(u)v=-v''+3u^2v.
\]

Now, we present the main convergence result with explicit constants by verifying the conditions of \Cref{thm:main}.

\begin{Theorem}[Main convergence theorem with explicit constants]\label{thm:mainPDE}
Let $v_0=0$ and $\lambda>0$ be the Tikhonov parameter. Assume that the iterates
$\{v_k\}$ remain in \(B_r := \{v\in \V : \|v\|_\V \le r\}\) for some $r>0$.
Define 
\[
\eta:=\|\F'(0)^{-1}\F(0)\|_{\V}=\|f\|_{H^{-1}},\qquad
L=\frac{3}{2\pi^2}r,\qquad
M=1+\frac{3}{4\pi^2}r^2.
\]
Then all the conclusions of \Cref{thm:main} hold true for this problem. In particular,
the Lipschitz constant in \Cref{thm:main} can be chosen as
\[
\widetilde L\ :=\ \frac{L}{1-\bar\theta}
\ \ =\ \frac{\frac{3}{2\pi^2}r}{\,1-\frac{\lambda}{1+\lambda}-M^2\varepsilon_\lambda \,}\,,
\]
where $\varepsilon_\lambda$ is the design/learning error,
and $\bar\theta:=\frac{\lambda}{1+\lambda}+M^2\varepsilon_\lambda$
bounds the forcing terms $\theta_k$ in \eqref{eq:forcing_varlam} for all $v_k$ with $v_k\in B_r$.
\end{Theorem}

\section{Conclusion and Outlook} \label{sec:conclusion_outlook}

In this paper, we introduced \textsc{CHONKNORIS}, a novel operator learning methodology rooted in the Newton--Kantorovich method for solving infinite-dimensional systems. Crucially, \textsc{CHONKNORIS} removes the longstanding ceiling in accuracy typically observed in operator learning frameworks, setting a new standard in precision. To our knowledge, this is the first instance of an operator learning framework that achieves machine precision in solving a variety of nonlinear partial differential equations (PDEs) and inverse problems. 

The central innovation of \textsc{CHONKNORIS} lies in employing an iterative scheme based on the Newton--Kantorovich method, where we explicitly learn the dependence of the Cholesky factors of the matrix $(\lambda I + (\frac{\partial \mathcal{F}}{\partial v})^*\frac{\partial \mathcal{F}}{\partial v})^{-1}$ on coefficients $u$ and solution estimates $v$. The explicit integration of the underlying equation within these iterations significantly enhances the accuracy and interpretability of the learned solutions. It also makes it uniquely suited for inverse problems, as the general equation $\mathcal{F}$ may incorporate observed data. The ability to tackle inverse problems directly within the operator learning framework is a significant improvement from traditional approaches, which depend on repeated iterations of forward solvers and often struggle in high-dimensional settings. Building on \textsc{CHONKNORIS}, we introduced \textsc{FONKNORIS}, a foundational model motivated by the observation that $\frac{\partial \mathcal{F}}{\partial v}$ always defines a linear PDE. While \textsc{CHONKNORIS} learns to solve the linearized PDEs associated with the Fr\'{e}chet derivative of one equation $\mathcal{F}$, \textsc{FONKNORIS} extends this concept by learning to solve any linearized PDE. By plugging this learned linear PDE solver into the \textsc{CHONKNORIS} framework, we obtain a universal operator learner capable of solving a broad class of nonlinear PDEs.

We validated our methodology on diverse forward problems, including a nonlinear elliptic equation, Burgers' equation, a Darcy flow equation, the Sine--Gordon equation, and Klein--Gordon equation as well as inverse problems, including Calder\`{o}n's problem, an inverse wave scattering problem, and a problem from seismic imaging full waveform inversion. Both \textsc{CHONKNORIS} and \textsc{FONKNORIS} achieve machine precision in these cases, demonstrating their robustness and accuracy. Notably, \textsc{FONKNORIS} attained near machine precision on the Klein--Gordon and Sine--Gordon equations, despite not being trained on these equations, demonstrating remarkable generalization capabilities.

Despite these advances, our method is still constrained by the computational cost and occasional ill-conditioning that arise when computing and storing Cholesky factors in high dimensions. In practice, avoiding explicit computation and inversion of the Fr\'{e}chet derivative is crucial. For example, rather than forming the derivative exactly, one can use the approximation described in \Cref{se:approx.Frechet}. Additional improvements in high-dimensional scalability depend on the surrogate model used to solve the linear subproblem. We note that \textsc{CHONKNORIS} and \textsc{FONKNORIS} are agnostic to the underlying machine learning algorithm used to approximate the Cholesky factors, which may allow one to solve the computational challenges associated with high-dimensional problems. In particular, this adaptability enables the integration of various deep learning techniques, such as Fourier Neural Operators or Deep Operator Networks, as well as scalability enhancements for Gaussian process regression through inducing points or nested Kriging. In \Cref{se:min_max_ordering}, we show that a sparse-Cholesky strategy can scale and accelerate CHONKNORIS for Gaussian-process surrogates, albeit with an accuracy-speed trade-off: faster run-times come at the cost of reduced precision. Even though, we don't make use of the sparse-Cholesky algorithm, we do exploit the rank-revealing and sparsity properties of the max-min ordering, see \Cref{se:rank_revealing_property}. Further ways of scaling and accelerating CHONKNORIS are by using nested kriging and approximating the Fr\'{e}chet derivative, see \Cref{sec:nested_kriging} and \Cref{se:approx.Frechet}, respectively. We also experimented with learning Newton–-Kantorovich increments directly to reduce overhead; however, this has thus far yielded only limited gains. Finally, incorporating multi-fidelity, multi-scale, and hierarchical approaches offer promising directions for future research.

\section{Code and data availability}

An open source Python implementation of all experiments presented in this work is available at \url{https://github.com/ArasBacho/CHONKNORIS}. We primarily
utilize the Python packages PyTorch \cite{pytorch}, PyTorch Lightning \cite{Falcon_PyTorch_Lightning_2019}, GPyTorch \cite{gpytorch}, and FastGPs \cite{fastgps}. Our implementation uses double-precision floating point
arithmetic for which machine precision is $2.2 \times 10^{-16}$. 

All synthetic datasets used in the forward and inverse PDE experiments are provided in the repository. For the seismic imaging (full waveform inversion) experiments we use the publicly available OpenFWI dataset \cite{deng2022openfwi}.

\section{Acknowledgments.} 
AB, XY, MD, TB and HO acknowledge support from the Air Force Office of Scientific Research under MURI awards number FA9550-20-1-0358 (Machine Learning and Physics-Based Modeling and Simulation), FOA-AFRL-AFOSR-2023-0004 (Mathematics of Digital Twins), the Department of Energy under award number DE-SC0023163 (SEA-CROGS: Scalable, Efficient, and Accelerated Causal Reasoning Operators, Graphs and Spikes for Earth and Embedded Systems), the National Science Foundation under award number 2425909 (Discovering the Law of Stress Transfer and Earthquake Dynamics in a Fault Network using a Computational Graph Discovery Approach). HO acknowledges support from the DoD Vannevar Bush Faculty Fellowship Program under ONR award number N00014-18-1-2363.  
AH and BH acknowledge support from the National Science Foundation under awards 
2208535 (Machine Learning for Bayesian Inverse Problems) and 
2337678 (CAREER: Gaussian Processes for Scientific Machine
Learning: Theoretical Analysis and Computational Algorithms). AH acknowledges support from a Carl S. Pearson Fellowship. EC acknowledges support from the Department of Defense (DoD) Vannevar Bush Faculty Fellowship held by Prof. Andrew Stuart (award N00014-22-1-2790), and the Resnick Sustainability Institute. AS acknowledges that material is based upon work supported by the U.S. Department of Energy, Office of Science, Office of Workforce Development for Teachers and Scientists, Office of Science Graduate Student Research (SCGSR) program. The SCGSR program is administered by the Oak Ridge Institute for Science and Education for the DOE under contract number DE-SC0014664.

\section{Authors contribution}

Conceptualization: HO, BH, AB, AS, XY; Methodology: HO, BH, AB, AS, AH, XY; Software: AB, AS, XY; Validation: AB, AS, XY, EC, MD; Formal analysis: HO, BH, AB, AS, XY; Investigation: AB, AS, XY; Writing -- Original Draft: AB, AS, TB, EC, BH, HO; Visualization: AB, AS, XY; Supervision: BH, HO; Funding acquisition: BH, HO; Writing -- Review \& Editing: All authors.


 \bibliographystyle{plain}
 \bibliography{references}

\appendix

\section{Gaussian process benchmarks} \label{sec:gp_benchmarks}

The Gaussian process (GP) benchmarks in \Cref{tab:summary-perf} are based on \cite{batlle2024kernel}. The GP measurements are given by either pointwise measurements or projection onto PCA coefficients, and the intermediate vector valued map is learned using a Gaussian process with a linear combination of a Mat\'{e}rn and dot product kernels. All hyper-parameters (number of PCA coefficients, lengthscale, regularization...) are learned via 5 fold cross validation on the training set and optimized using Optuna \cite{optuna}. 

\section{Nested Kriging} \label{sec:nested_kriging}
Nested Kriging is a hierarchical extension of Gaussian process regression designed to handle very large datasets that cannot be represented efficiently by a single GP. The idea is to train separate specialized GP models (experts) on subsets of the data or related problems, and then combine (aggregate) their predictions using another GP or a statistically optimal weighting scheme. This aggregation step, often based on minimizing the overall prediction variance, produces a global predictor that retains much of the accuracy of a full GP model while drastically reducing computational and memory costs. 

More formally, let $M_1(x), M_2(x), \dots, M_p(x) \in \mathbb{R}^{m}, x\in \mathbb{R}^n$ be a set of $p\in \mathbb{N}$ experts that are described by the same underlying Gaussian process $Y(x)$ with kernel $k$, i.e., there holds
\begin{align*}
    M_i(x)=k(x,X_i)k(X_i,X_i)^{-1}Y(X_i), \quad i=1,2,\dots, p.
\end{align*}

Furthermore, define $M(x)=(M_1(x),M_2(x), \dots, M_p(x))^{T}$ as well as the covariance matrix $K_M(X)=\mathrm{Cov}(M(x),M(x))$ and the vector $k_M(x)=\mathrm{Cov}(Y(x),M(x))$. Then, the aim is to minimize the variance 
\begin{align*}
E[(Y(x) - \alpha^T M(x))^2]= k(x, x) - 2 \alpha^T k_M(x) + \alpha^T K_M(x) \alpha
\end{align*}
with respect to the aggregation vector $\alpha$. The solution to this mean squared problem is given by the 
\begin{align*}
\alpha^* = K_M(x)^{-1} k_M(x)
\end{align*}
leading to the unconstrained best linear unbiased predictor (BLUP) given by 
\begin{align*}
M_{\mathcal{A}}(x) = K_M(x)^{-1} k_M(x) M(x). 
\end{align*} However, we aim to aggregate our models by weighting the results of the individual experts by incorporating the constraint $\sum_{i \in A} \alpha_i(x) = \mathbf{1}^T \alpha(x)= 1$ for all $x\in \mathbb{R}^n, $ where \( \mathbf{1} \) is a vector of ones with the same dimension as \( \alpha \). The constrained optimization problem is then solved via Lagrange multipliers. The corresponding Lagrange function is given by
\begin{align} \label{eq:model.agg}
    \begin{split}
        \mathcal{L}(\alpha, \lambda)
 &= E[(Y(x) - \alpha^T M(x))^2] + \lambda (\mathbf{1}^T \alpha - 1)\\
    &=k(x, x) - 2 \alpha^T k_M(x) + \alpha^T K_M(x) \alpha + \lambda (\mathbf{1}^T \alpha - 1)
    \end{split}
\end{align}
where \( \lambda \) is the Lagrange multiplier associated with the constraint \( \mathbf{1}^T \alpha = 1 \). This gives the following result proved in \Cref{sec:appendix_proofs}

\begin{Proposition} 
\label{prop:lagrange_func_min} 
The Lagrange function given by \eqref{eq:model.agg} is minimized by
\begin{align}\label{eq:alpha}
 \alpha = K_M(x)^{-1} \left( k_M(x) - \frac{\mathbf{1} \left( \mathbf{1}^T K_M(x)^{-1} k_M(x) - 1 \right)}{\mathbf{1}^T K_M(x)^{-1} \mathbf{1}} \right).
\end{align} 
\end{Proposition}

\begin{proof}[Proof of \Cref{prop:lagrange_func_min}]
To minimize \( \mathcal{L}(\alpha, \lambda) \), we take the derivative of \( \mathcal{L} \) with respect to \( \alpha \) and \( \lambda \), and set them equal to zero. This gives the equations:
\begin{align*}
    K_M(x) \alpha = k_M(x) - \frac{\lambda}{2} \mathbf{1}, \quad \mathbf{1}^T \alpha = 1.
\end{align*}
Substituting the $\alpha$ in the first equation into the second equation, solving for $\lambda$, and substituting $\lambda$ back into the first equation gives the new values of \( \alpha \) that satisfy the constraint \( \sum_{i \in A} \alpha_i(x) = 1 \).
This finally yields the desired formula \eqref{eq:alpha}.
\end{proof}

By enforcing the constraint \( \sum_{i \in A} \alpha_i(x) = 1 \), the coefficients \( \alpha_i(x) \) are adjusted from the unconstrained BLUP solution, and the Lagrange multiplier method gives us the corrected formula. This ensures that the coefficients sum to 1 while still minimizing the mean squared error in the best linear unbiased way.

\subsubsection*{Computational complexity}  
The computational cost of the Nested Kriging can be divided into two parts: the offline training of the individual experts and the online prediction (aggregation).  
During training, each expert \( M_i \) requires the inversion (or Cholesky factorization) of its covariance matrix \( K_i \in \mathbb{R}^{n_i \times n_i} \), resulting in a computational cost of order \( \mathcal{O}(n_i^3) \) and a memory requirement of \( \mathcal{O}(n_i^2) \). Training all \( p \) experts therefore costs \( \mathcal{O}\!\left(\sum_{i=1}^p n_i^3\right) \), which is significantly cheaper than training a single global Gaussian process on all \( N = \sum_{i=1}^p n_i \) data points, whose cost would be \( \mathcal{O}(N^3) \). For experts of equal size (\( n_i = n \)), this corresponds to a reduction by a factor of approximately \( p^2 \).  

At prediction time, computing the prediction \( M_{\mathcal{A}}(x) \) for a new input \( x \) involves two main steps. First, for each expert, we compute the predictive weights \( v_i(x) = K_i^{-1} k(X_i, x) \), which requires \(\mathcal{O}(n_i^2)\) operations per expert, yielding a total cost of \(\mathcal{O}\!\left(\sum_{i=1}^p n_i^2\right)\). These vectors are then used to construct the aggregated covariance terms: \( k_M(x) \in \mathbb{R}^p \) and \( K_M(x) \in \mathbb{R}^{p \times p} \), where forming all pairwise covariances typically costs \(\mathcal{O}(N^2)\) if cross-covariances between experts are precomputed. Finally, the aggregation weights \(\alpha(x)\) are obtained by inverting \(K_M(x)\), which adds an additional \(\mathcal{O}(p^3)\) cost. Consequently, the total online prediction complexity per test point is of order \(\mathcal{O}(N^2 + p^3)\), compared to \(\mathcal{O}(N^2)\) for a single global Gaussian process.  

Hence, Nested Kriging achieves a substantial reduction in offline training cost—from cubic in the total dataset size to the sum of the individual cubic costs—while maintaining comparable prediction complexity and offering improved scalability with respect to both memory and computation.

\section{Choice of ordering in the Cholesky Factorization} \label{se:min_max_ordering}
The Cholesky factors of $\mathcal{Q}$   depend on the ordering of its rows and columns, which corresponds to the degrees of freedom in $\mathbb{R}^N$. While this ordering can be inherited from the discretization of the space $\mathcal{V}$, the accuracy of the approximation in \eqref{eq:cholesky_approximation} can be significantly improved by selecting an ordering that induces a hierarchical or multiresolution structure in the operator problem  \cite{owhadi2017multigrid, owhadi2019operator}.
To describe this, consider first the case where $\mathcal{F}$ is a local differential operator (e.g., a PDE) acting on functions defined over a domain $\Omega \subset \mathbb{R}^d$. Here,
$\frac{\delta \mathcal{F}}{\delta v}^* \frac{\delta \mathcal{F}}{\delta v} + \lambda I$
can be interpreted as a (discretized) elliptic PDE, and its inverse, $\Theta$, represents a (discretized) Green’s function.
Suppose the discretization of $\mathcal{F}$ is obtained using a numerical method (e.g., finite element, finite difference, or collocation), with elements centered around points $x_1, \ldots, x_N \in \Omega$. Although the Cholesky factors of $\Theta$ are generally dense when using a lexicographic ordering, they exhibit significant sparsity when a \emph{max-min ordering} is used. As presented in \cite{schafer2021compression}, a max-min ordering $\pi$ of $\{1, \ldots, N\}$ is defined such that:
\begin{equation*}
\pi(1) = \operatorname{argmax}_i \operatorname{dist}(x_i, \partial \Omega)\,,
\end{equation*}
and for $i \geq 1$:
\begin{equation*}
\pi(i+1) = \operatorname{argmax}_j \operatorname{dist}\big(x_j, \partial \Omega \cup \{x_{\pi(1)}, \ldots, x_{\pi(i)}\}\big)\,.
\end{equation*}
Writing $\Pi$ for the permutation matrix associated with $\pi$ and
 \begin{equation*}
 \Pi^T L^\dagger L^{\dagger,T} \Pi= \Theta
 \end{equation*}
  for the exact Cholesky factorization of $\Theta$ in the maxmin ordering,  those Cholesky factors have two desirable properties \cite{schafer2021compression, schafer2021sparse, chen2024sparse}: (1) they are rank revealing, and (2) they are sparse.
  
\subsubsection{Rank-Revealing Property}
The matrix $L^{\dagger,(k)}$, defined as the truncation of $L^\dagger$ to its first $k$ columns (with all remaining columns set to zero), provides a near-optimal rank-$k$ approximation (measured in operator norm) within a constant factor \cite[Thm.~2.3]{schafer2021compression}. Specifically, it satisfies:
\begin{equation*}
\big\|\Theta - \Pi^T L^{\dagger,(k)} L^{\dagger,(k),T} \big\|_{\mathrm{Fro}} \lesssim \inf_{M \text{ of rank } k} \big\|\Theta - M \big\|_{\mathrm{Fro}},
\end{equation*}
where $\|\cdot\|_{\mathrm{Fro}}$ denotes the Frobenius matrix norm.
This rank-revealing property is: (a) Analogous to the properties achieved with gamblets or operator-adapted wavelets \cite{owhadi2017multigrid, owhadi2019operator} for the compression of elliptic PDEs. (b)  more desirable for operator-learning than mesh invariance because it induces, within a constant factor, an optimal approximation of the underlying operator at any level of truncation/discretization which has been the core objective of numerical homogenization \cite{altmann2021numerical}.

\subsubsection{Sparsity}
The Cholesky factors $L^\dagger$ exhibit exponential decay in their entries, as shown in \cite[Thm.~5.23]{schafer2021compression}. Specifically:
\begin{equation*}
|L^\dagger_{i,j}| \lesssim \exp\left(- C \frac{\operatorname{dist}(x_{\pi(i)}, x_{\pi(j)})}{h}\right)\,,
\end{equation*}
where $h$ is the mesh norm associated with $\{x_1, \ldots, x_N\}$, and $C$ is some positive constant.
To describe the sparsity structure, let $l_1 := \max_i \operatorname{dist}(x_i, \partial \Omega)$ and, for $i \geq 1$,
\begin{equation*}
l_{i+1} := \max_j \operatorname{dist}\big(x_j, \partial \Omega \cup \{x_{\pi(1)}, \ldots, x_{\pi(i)}\}\big)\,,
\end{equation*}
represent the sequence of decreasing distances associated with the max-min ordering. Given $\rho \in \mathbb{N}$, the sparsity set is defined as:
\begin{equation*}
S_\rho := \big\{(i,j) \in \{1, \ldots, N\}^2 \mid i \geq j \text{ and } \operatorname{dist}(x_{\pi(i)}, x_{\pi(j)}) \leq \rho l_i \big\}.
\end{equation*}
The set $S_\rho$ contains approximately $\mathcal{O}(N \rho^d)$ elements. Using an incomplete Cholesky factorization with sparsity pattern $S_\rho$, one obtains a lower triangular matrix $L_\rho$ with non-zero entries restricted to $S_\rho$. This approach achieves the approximation:
\begin{equation*}
\big\| \Pi^T L_\rho L_\rho^T \Pi - \Theta \big\|_{\mathrm{Fro}} \lesssim e^{-C \rho}.
\end{equation*}
Consequently, it is possible to achieve an accuracy $\epsilon$ by accessing only $\mathcal{O}(N \log^d \frac{1}{\epsilon})$ entries of $\Theta$. The resulting Cholesky factors $L_\rho$ will have just $\mathcal{O}(N \log^d \frac{1}{\epsilon})$ non-zero entries.

\subsubsection{Exploiting the maxmin ordering for operator learning} \label{se:rank_revealing_property}
The rank revealing and sparsity properties of the maxmin ordering have been leveraged in \cite{SchOwh24SRES} for achieving SOTA in terms of both complexity and data efficiency for learning the solution operator of 
arbitrary elliptic PDEs (  \cite{SchOwh24SRES} shows that those solution operators can be rigorously approximated to accuracy $\epsilon$ from only $\mathcal{O}(\log N \log^d\frac{N}{\epsilon})$ source-solution pairs).
Here we employ the maxmin ordering to improve the accuracy of the map $v\rightarrow L(v)$ such that
\begin{equation}\label{eq:cholesky_approximation2}
\Pi^T L L^T \Pi \approx \Theta\,. 
\end{equation}

\subsubsection{Exploiting Random Ordering for Operator Learning}
The rank-revealing and sparsity properties of the max-min ordering can also be achieved using a random ordering, as it implicitly induces a max-min structure (see \cite[Thm.~2.9]{owhadi2017multigrid}, derived from \cite{halko2011finding}). 
In cases where $\mathcal{F}$ is high-dimensional or represents an arbitrary operator not necessarily associated with a PDE, the rank-revealing property is still preserved by the random ordering 
\cite{chen2023randomly}, therefore in those settings we can simply replace $\Pi$ in \eqref{eq:cholesky_approximation2} by random permutation matrix.

 \section{Discretizations and Cholesky factorization}
In practice, we work with an arbitrary discretization of the operator $\mathcal{G}$, setting $\mathcal{U}= \mathbb{R}^J$ and $ \mathcal{V} = \mathbb{R}^K$. This discretization can be derived from various numerical methods such as finite-element, or finite-difference methods or any other discretization method such as the optimal recovery approach in \cite{batlle2024kernel}.
For simplicity, we retain the original notation for the discretized operators, continuing to use $\mathcal{G}$ and $\mathcal{F}$ to denote their discretized counterparts. The spaces $\mathcal{U} $ and $\mathcal{V}$ are equipped with the Euclidean inner product. Consequently, the derivative $\frac{\delta \mathcal{F}}{\delta v}$ becomes a function mapping $\mathbb{R}^J$ to the space of $K \times K$ matrices.
The elliptic operator
\begin{equation*}
\frac{\delta \mathcal{F}}{\delta v}^* \frac{\delta \mathcal{F}}{\delta v} + \lambda I \,:\, \mathcal{V} \to \mathcal{L}_+(\mathcal{V})
\end{equation*}
then simplifies to a map from $\mathbb{R}^J$ to the space of $J \times J$ symmetric positive definite matrices. Then learning the operator $\mathcal{Q}$ reduces to learning a mapping from $\mathbb{R}^J$ to the space of $J \times J$ symmetric positive definite matrices. To facilitate this, we represent $\mathcal{Q}$ through its Cholesky factorization:
\begin{equation*}
\mathcal{Q} = \mathcal{R} \mathcal{R}^T\,,
\end{equation*}
and seek to learn a lower-triangular-matrix-valued operator:
\begin{equation*}
\mathcal{R} : \mathcal{V} \rightarrow \{R \in \mathbb{R}^{J \times J} \mid R \text{ is lower triangular}\},
\end{equation*}

Then a neural network or matrix valued RKHS can be used to parameterize 
and learn $\mathcal{R}$. In practice we find that this parameterization of 
the DNO leads  to higher accuracy as the composed operators will enforce positivity and symmetry of $\mathcal{Q}$, however, one can design intermediate algorithms 
where $\mathcal{Q}$ or $\mathcal{N}$ are directly learned.

\section{Choice of the Learning rate $\alpha$ and Relaxation $\lambda$} \label{se:Learning_rate_relaxation}
\begin{figure}[!ht]
\centering
    \begin{tikzpicture}
        \draw (0,0) node[draw,circle](zz){$\lambda_n,\alpha_n$};
        \draw (-2.5,0) node[draw,circle](nz){$\kappa_\lambda \lambda_n,\alpha_n$};
        \draw (-2.5,2.5) node[draw,circle](np){$\kappa_\lambda \lambda_n,\beta_\alpha \alpha_n$};
        \draw (-2.5,-2.5) node[draw,circle](nn){$\kappa_\lambda \lambda_n,\kappa_\alpha \alpha_n$};
        \draw (0,2.5) node[draw,circle](zp){$\lambda_n,\beta_\alpha \alpha_n$};
        \draw (0,-2.5) node[draw,circle](zn){$\lambda_n,\kappa_\alpha \alpha_n$};
        \draw (2.5,0) node[draw,circle](pz){$\beta_\lambda \lambda_n,\alpha_n$};
        \draw (2.5,2.5) node[draw,circle](pp){$\beta_\lambda \lambda_n,\beta_\alpha \alpha_n$};
        \draw (2.5,-2.5) node[draw,circle](pn){$\beta_\lambda \lambda_n,\kappa_\alpha \alpha_n$};
        \draw[-] (zz) -- (nn); 
        \draw[-] (zz) -- (nz); 
        \draw[-] (zz) -- (np); 
        \draw[-] (zz) -- (zn); 
        \draw[-] (zz) -- (zp); 
        \draw[-] (zz) -- (pn); 
        \draw[-] (zz) -- (pz); 
        \draw[-] (zz) -- (pp); 
    \end{tikzpicture}
    \caption{Options for jointly adaptive relaxation and learning rate.}
    \label{fig:adaptive_relax_lr_fwi}
\end{figure}
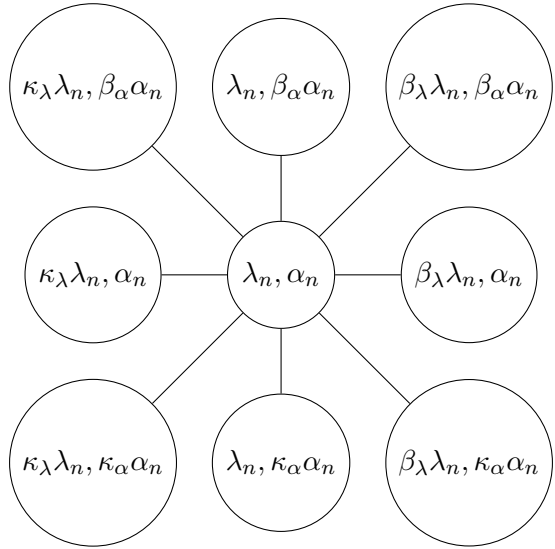

We often also found it necessary to adapt the relaxation $\lambda$ across iterations for convergence to machine precision in a reasonable number of steps. Generally speaking, $\lambda$ is decreased as our NK/\textsc{CHONKNORIS} approximation approaches the true solution in order to achieve the quadratic convergence of Gauss--Newton steps. In such scenarios, the learned Cholesky factor $\hat{\mathcal{R}}$ has an additional dependence on the relaxation $\lambda$. To simultaneously tune the learning rate and relaxation, at any given iteration we choose to inflate, deflate, or keep both parameters from the previous iteration. Specifically, if $(\lambda_n,\alpha_n)$ are the values at iteration $n$, then we choose $\lambda_{n+1} \in \{\beta_\lambda \lambda_n, \lambda_n , \kappa_\lambda \lambda_n \}$ and $\alpha_{n+1} \in \{\beta_\alpha \alpha_n , \alpha_n, \kappa_\alpha \alpha_n\}$ among all $9$ possible combinations where $\kappa_\lambda,\kappa_\alpha \in (0,1)$ are decay factors and $\beta_\lambda, \beta_\alpha$ are inflation factors.  
The choices are visualized in \Cref{fig:adaptive_relax_lr_fwi}.   

\section{Approximation of the Fr\'{e}chet derivative}\label{se:approx.Frechet}
 In this section, we discuss some numerical approximations of the Jacobian in cases when the explicit computation and storage of the Jacobian becomes infeasible, e.g., for high-dimensional problems such as problems in seismic imaging cf. \Cref{sec:seismic_imaging}. The idea is to approximate the Fr\'{e}chet derivative of $\mathcal{F}$ by the finite difference 
 \begin{align*}
\frac{\delta \mathcal{F}}{\delta v}(u,v)[h] \approx   \frac{\mathcal{F}(u,v+t h)-\mathcal{F}(u,v)}{t},
 \end{align*} 
 for sufficiently small $t>0$ justified by the limit
 \begin{align*}
\frac{\delta \mathcal{F}}{\delta v}(u,v)[h] =\lim_{t\searrow 0}  \frac{\mathcal{F}(u,v+t h)-\mathcal{F}(u,v)}{t}.
 \end{align*} We obtain for all $h\in \mathcal{V}$
\begin{align}\label{eq.approx.jacobian}
\left \langle \left(\frac{\delta \mathcal{F}}{\delta v}(u,v)\right)^* \mathcal{F}(u,v), h \right \rangle_{\mathcal{V}} = \left \langle \mathcal{F}(u,v), \frac{\delta \mathcal{F}}{\delta v}(u,v) h\right\rangle_{\mathcal{W}} \approx \frac{1}{t}\left \langle \mathcal{F}(u,v), \mathcal{F}(u,v+t h)-\mathcal{F}(u,v) \right\rangle_{\mathcal{W}}
\end{align}
Denoting again with $\mathbb{R}^{N_{\mathcal{V}}}$ the finite-dimensional approximation of the space $\mathcal{V}$, we can find the values $\frac{\delta \mathcal{F}}{\delta v}^* \mathcal{F}(u,v)$ by choosing $h$ to be the values of a basis $\mathcal{B}$ of $\mathbb{R}^{N_{\mathcal{V}}}$. In order to further reduce the complexity, we can perform a singular value decomposition of  $\frac{\delta \mathcal{F}}{\delta v}^* \mathcal{F}(u,v):\mathcal{U}\times \mathcal{V}\rightarrow \mathcal{V}$ on a given dataset $\lbrace (u_i,v_i)\rbrace_{i=1}^m \subset \mathcal{U}\times \mathcal{V} $: Denoting with $F\in \mathbb{R}^{m\times N_{\mathcal{V}}}$ the matrix collocating the point evaluations $\frac{\delta \mathcal{F}}{\delta v}^* \mathcal{F}(u_i,v_i)$, we obtain $F=U \Sigma V^T$ with $U\in \mathbb{R}^{m\times m}, \Sigma \in  \mathbb{R}^{m\times N_{\mathcal{V}}},$ and $V\in  \mathbb{R}^{N_{\mathcal{V}}\times N_{\mathcal{V}}}$. Given a low-rank approximation $r<< N_{\mathcal{V}}$, we have $F\approx U_r \Sigma_r V_r^T$. Then, an approximation of $\left(\frac{\delta \mathcal{F}}{\delta v}(u,v)\right)^* \mathcal{F}(u,v)$ can be achieved by expanding it in terms of the reduced basis $\mathcal{B}_r=\lbrace v_1, \dots, v_r\rbrace$ where the coefficients of this can be found by testing with $h=v_i, i=1,\dots,r$ in $\eqref{eq.approx.jacobian}$.

\section{Connection to Attention Mechanism}
\label{se:attention}
The \textit{operator-valued operator} $\mathcal{N}(v_n)=\left(\left(\frac{\delta \mathcal{F}}{\delta v}(u,v_n)\right)^* \frac{\delta \mathcal{F}}{\delta v}(u,v_n) + \lambda I\right)^{-1}$ depends nonlinearly on its input (the current iterate $v_n$), and its output is a nonlocal, linear operator. This observation leads to the connection with the attention mechanism proposed in \cite{vaswani2017attention}, which defines itself an operator of this nature. The attention mechanism is at the heart of transformers, a neural network architecture widely used for language \citep{touvron_llama_2023} and vision \citep{dosovitskyi2021image} tasks and recently employed in the context of operator learning, for example as in \cite{cao2021choose, hao2023gnot,li2023transformer,rahman2024pretraining,Calvello2024Continuum}. Indeed, self-attention and cross-attention are extended to the function space setting in \cite{Calvello2024Continuum}. Following this framework, we write the extension to function space of the attention mechanism from \cite{vaswani2017attention} in its most general form.  
We define the attention operator $\mathcal{A}: \mathcal{V}\times \mathcal{W}\times \mathcal{U} \to \mathcal{U}$ by
\begin{equation}
\label{eq:attention_operator}
\bigl(\mathcal{A}(v,w,u) \bigr)(x)=  \mathbb{E}_{y\sim p(\cdot;v,w,x)}Vu(y),   
\end{equation}
where 
\begin{equation*}
    p(y;v,w,x) = \frac{\mathrm{exp} \Big ( \big\langle Q v(x), K w(y) \big\rangle \Big )}{\int_{\Omega} \mathrm{exp} \Big ( \big\langle Q v(x), K w(s) \big\rangle \Big ) \: \mathrm{d}s},
\end{equation*}
where the learnable $Q \in \mathbb{R}^{d\times d_v},K\in \mathbb{R}^{d\times d_w},V\in \mathbb{R}^{d\times d_u}$ parametrize the attention operator $\mathcal{A}$. Given this definition, we may approximate \begin{equation}
\label{eq:attention_approx}
    \mathcal{N}(v_n)\approx \mathcal{A}(v_n,v_n,\cdot) \in \mathcal{L}(\mathcal{U}, \mathcal{V}).
\end{equation}
The above is unlike the standard way attention is implemented in practice, as the second and third input functions of $\mathcal{A}$ differ. An alternative approximation of $\mathcal{N}$ based on the cross-attention operator is given by $\mathcal{A}(v_n, \mathcal{F}(u,v_n),\cdot)$; note that, here, the second input is the residual itself. 
Using an approximation of $\mathcal{N}$ given by \eqref{eq:attention_approx} leads to the reformulation of the iteration \eqref{eq:RegNK.increment} as a neural network block\begin{equation}
\label{eq:chonkNet_block}
    \left\{
    \begin{matrix}
        r_{n+1} &=& \mathcal{F}(u,v_n),\\
        \delta v_{n+1} &=& \mathcal{A}(v_n,v_n,r_{n+1})\approx \mathcal{N}(v_n)(r_{n+1}),\\
        v_{n+1} &=& v_n + \delta v_{n+1}.
    \end{matrix}
    \right. 
\end{equation}
We highlight that as the transformer from \cite{vaswani2017attention} consists of the application of the attention mechanism (a nonlocal operator) along with residual connections and pointwise linear transformations, the block in \eqref{eq:chonkNet_block} is defined itself by a pointwise transformation, attention and a residual connection. This perspective may be used as the basis for an end-to-end learning approach. In particular, we may unroll \textsc{CHONKNORIS}, as outlined in \cite{monga2021algorithm}. This entails choosing a fixed number of iterations for our iterative method and interpreting this iterative method as a composition of blocks of the form \eqref{eq:chonkNet_block}. We leave further investigation of this end-to-end approach to future work.

\begin{Remark}
Note that while common operator learning models fail to achieve high  accuracy with increasing depth, the proposed iterative structure \eqref{eq:RegNK.increment} is specifically designed to converge toward the true operator $\mathcal{G}$ as depth increases
as the compositions emulate a quasi-Newton algorithm.
This connection aligns with observations in the ANN literature \cite{monga2021algorithm, owhadi2023ideas}, where successful algorithms are effective not solely due to their depth or complexity, but because they possess: (1) Sufficient expressivity to emulate convergent numerical approximation methods; and (2) Enough structure in their computational graphs \cite{owhadi2022computational}
 to leverage the same principles that make such numerical methods effective. Without (2) increasing depth in an ANN algorithm may only increase expressivity and complexity without improving accuracy.
\end{Remark}

\section{Proofs} 
\label{sec:appendix_proofs}

\subsection{Theoretical results}

\begin{proof}[Proof of \Cref{thm:main}]
Write $A(v):=\F'(v)$, $A_k:=A(v_k)$, and $F_k:=\F(v_k)$.
For $\lambda>0$, $R_\lambda(v):=(\lambda I_{\mathcal V}+A(v)^*A(v))^{-1}$.

\emph{Step 1.}
On $\mathcal W$,
\begin{equation}\label{eq:tikh_id_varlam}
I_{\mathcal W}-A(v)\,R_\lambda(v)\,A(v)^* \;=\; \lambda\big(\lambda I_{\mathcal W}+A(v)A(v)^*\big)^{-1}.
\end{equation}
Indeed, $(\lambda I_{\mathcal V}+A^*A)R_\lambda=I_{\mathcal V}$; multiplying on the left by $A$ and on the right by $A^*$ gives
$A(\lambda I + A^*A)R_\lambda A^* = AA^*$, i.e.,
$\lambda A R_\lambda A^* + AA^* R_\lambda A^* = AA^*$,
and rearranging yields \eqref{eq:tikh_id_varlam} since $(\lambda I_{\mathcal W}+AA^*)$ is invertible by (A3).

\emph{Step 2.}
With $\delta v_k=-\widehat R_{\lambda_k}(v_k)A_k^*F_k$,
\[
A_k \delta v_k + F_k
= \big(I - A_k \widehat R_{\lambda_k}(v_k) A_k^*\big)F_k
= \underbrace{\big(I - A_k R_{\lambda_k}(v_k) A_k^*\big)}_{\text{(I)}}F_k
\ +\ \underbrace{A_k\big(R_{\lambda_k}(v_k)-\widehat R_{\lambda_k}(v_k)\big)A_k^*}_{\text{(II)}}F_k.
\]
By \eqref{eq:tikh_id_varlam} at $v_k$ and (A3),
\[
\|(I)\|=\big\|\lambda_k(\lambda_k I_{\mathcal W}+A_kA_k^*)^{-1}\big\|
\le \frac{\lambda_k}{\lambda_k+\sigma_*^2} \;=:\; \theta^{\rm tik}_k.
\]
For (II), using submultiplicativity and \Cref{ass:NK}(A3),
\[
\|(II)\|\ \le\ \|A_k\|\,\|R_{\lambda_k}(v_k)-\widehat R_{\lambda_k}(v_k)\|\,\|A_k^*\|
\ \le\ M^2\,\varepsilon_{\lambda_k} \;=:\; \theta^{\rm des}_k.
\]
Therefore
\[
\|A_k \delta v_k + F_k\|
\ \le\ (\theta^{\rm tik}_k+\theta^{\rm des}_k)\,\|F_k\|
\ =:\ \theta_k\,\|F_k\|,
\]
which is the inexact Newton forcing inequality \eqref{eq:forcing_varlam}.

\emph{Step 3.}
By (A2), for any $v$ and $s$ with $v,v+s\in\overline{B(v_0,R)}$,
\[
\|\F(v+s)-\F(v)-\F'(v)s\|
\ \le\ \tfrac12 L \|s\|^2.
\]
The standard (Dembo--Eisenstat--Steihaug) majorant analysis applies with
\[
\bar\theta:=\sup_k \theta_k \;<\; 1,
\qquad
\widetilde L:=\frac{L}{1-\bar\theta},
\]
yielding existence/uniqueness in the ball, the majorant bounds, and convergence provided $\widetilde h=\beta \widetilde L \eta\le \tfrac12$ and $t_* \le R$.
If additionally $\lambda_k\to 0$ and $\varepsilon_{\lambda_k}\to 0$, then $\theta_k\to 0$ and the local rate approaches quadratic; if $\theta_k=\mathcal O(\|F_k\|)$, the rate is quadratic.
\end{proof}

\begin{proof}[Proof of \Cref{cor:rates1}]
We sketch the standard derivation; see also DES \cite{steighaug1982inexact}. By the mean-value theorem for G\^{a}teaux differentiable functions and (A2), we obtain
\[
\F(v_k+\delta v_k)-\F(v_k)-A(v_k)\delta v_k \ =\ \int_0^1\big(A(v_k+\tau \delta v_k)-A(v_k)\big)\,\delta v_k\,d\tau,
\]
so $\|\F(v_{k+1})\|\le \|A(v_k)\delta v_k+\F(v_k)\|+\tfrac{L}{2}\|\delta v_k\|^2$.
Left-multiplying by $A(v_k)^{-1}$ and using $\|A(v_k)^{-1}\|\le \beta$ on $\overline{B}(v_0,R)$ yields
\[
\|v_{k+1}-v_k-A(v_k)^{-1}\F(v_k)\|
\ \le\ \beta\,\|A(v_k)\delta v_k+\F(v_k)\|\ +\ \frac{\beta L}{2}\,\|\delta v_k\|^2.
\]
By the forcing condition with $\bar\theta$ and the stability of the step $\|\delta v_k\|\le \|A(v_k)^{-1}\|\,\|\F(v_k)\|+o(\|\F(v_k)\|)\le \beta \|\F(v_k)\|+o(\|\F(v_k)\|)$, one arrives at
\[
\|v_{k+1}-v^*\|\ \le\ \frac{\bar\theta}{1-\bar\theta}\,\|v_k-v^*\| \ +\ \frac{\beta L}{2(1-\bar\theta)}\,\|v_k-v^*\|^2
\]
which is \eqref{eq:one-step1}.
Assertions (i)--(iv) follow immediately: (i) by dropping the quadratic term and applying limes superior; (ii) because $\theta_k\to 0$ eliminates the linear term; (iii) and (iv) use the local equivalence $\|\F(v_k)\|\asymp \|v_k-v_*\|$ (since $A(v_*)$ is continuous and invertible for large $k$) to replace $\|\F(v_k)\|^\alpha$ by $e_k^\alpha$.
\end{proof}

\begin{proof}[Proof of \Cref{prop:schedule}]
(a) is immediate from $\theta_{\rm tik,k}\to 0$ and $\theta_{\rm des,k}\to 0$.
For (b), $\theta_{\rm tik,k}\le \lambda_k/\sigma_*^2=\mathcal O(\|\F(v_k)\|)$ and $\theta_{\rm des,k}=M^2\,\varepsilon_{\lambda_k}=\mathcal O(\|\F(v_k)\|)$, hence $\theta_k=\mathcal O(\|\F(v_k)\|)$; apply \Cref{cor:rates1} (iv).
For (c), differentiate $\phi$ on $(0,\infty)$:
\[
\phi'(\lambda)=\frac{\sigma_*^2}{(\lambda+\sigma_*^2)^2}-2\,C\,\lambda^{-3}.
\]
Setting $\phi'(\lambda)=0$ yields $(\lambda+\sigma_*^2)^{-2}\sim 2 C\,\sigma_*^{-2}\lambda^{-3}$; for the coarse scaling one may drop the $+\sigma_*^2$ inside parentheses (or solve exactly), giving $\lambda^{3}\sim 3 C\,\sigma_*^2$ and the stated $\lambda_*$. This balances the two terms in $\bar\theta$; an annealing $\lambda_k\downarrow 0$ then ensures $\theta_k\to 0$ and, by \Cref{cor:rates1}, superlinear/quadratic rates.
\end{proof}

\begin{proof}[Proof of \Cref{thm:mainPDE}]
We verify the assumptions of \Cref{thm:main} for the PDE setting and
identify the constants appearing there.

\medskip
\emph{Step 1: Setting and choice of base point.}
For a fixed $f\in \W$, we suppress the dependence on \(f\) and write \(\F(u) = \F(u,f)=-\Delta u+\kappa u^3-f :\mathcal V\to\mathcal W\).
For simplicity, we choose the base point \(v_0 = 0\) and $\kappa=1$.

By definition of the residual map,
\[
  \F(0) = -f \in \W, 
  \qquad 
  \F'(0) = A:=-\Delta : \V \to \W.
\]
Recall that the \(H^{-1}\)-norm is defined by duality with respect to the
\(\V\)-inner product induced by \(A\), so that for all \(g\in \W=H^{-1}\),
\[
  \|g\|_{\W}
  \;=\;
  \|A^{-1}g\|_{\V}.
\]
Therefore
\[
  \eta
  \;:=\;
  \|\F'(0)^{-1}\F(0)\|_\V
  \;=\;
  \|A^{-1}(-f)\|_\V
  \;=\;
  \|f\|_{\W},
\]
which is the first identity in the statement.

\medskip
\emph{Step 2: Verification of (A1) and explicit \(\beta\).}
Assumption (A1) of \Cref{ass:NK} requires that
\(\F'(v_0)\) is invertible and \(\|\F'(v_0)^{-1}\|\le \beta\).
As just noted, \(\F'(0)=A\) is an isomorphism \(\V\to \V^*=\W\), so (A1) holds with
\[
  \beta = \|\F'(0)^{-1}\| = \|A^{-1}\| = 1,
\]
where the last equality follows from the way the \(H^{-1}\)-norm is induced by
the \(\V\)-inner product via \(A\).

\medskip
\emph{Step 3: Lipschitz bound (A2) and choice of \(L\).}
Let \(B_r := \{v\in \V : \|v\|_\V \le r\}\). 
Then, for each \(v\in X\), we have 
\[
  \F'(v)h
  \;=\; Ah + 3v^2h.
\]
Hence, for \(u,v\in B_r\),
\[
  \big(\F'(u)-\F'(v)\big)h
  = 3\,(u^2 - v^2)h = 3\,(u+v)(u-v)h.
\]
Using the one-dimensional Sobolev embedding 
and the Poincar\'{e} inequality on \(\mathbb{T}\), we obtain
\[
  \|w\|_{L^\infty} \;\le\; \frac{1}{\pi}\,\|w\|_\V,
  \qquad
  \|w\|_{L^2} \;\le\; \frac{1}{\pi}\,\|w\|_\V.
\]
Together with the Cauchy-Schwarz inequality, we obtain the bound
\[
  \|\,(u^2 - v^2)h\,\|_{H^{-1}}
  \;\le\; \frac{1}{2\pi^2}\,\big(\|u\|_\V + \|v\|_\V\big)\,\|u-v\|_\V\,\|h\|_\V.
\]
Since \(u,v\in B_r\), this yields
\[
  \|\F'(u)-\F'(v)\|
  \;\le\; \frac{3}{2\pi^2}\,r\,\|u-v\|_\V,
\]
so (A2) holds with
\[
  L := \frac{3}{2\pi^2}\,r.
\]

\medskip
\emph{Step 4: Uniform bound (A3) and choice of \(M\) and \(\sigma_*\).}
For any \(v\in B_r\) and \(h\in \V\) with \(\|h\|_\V=1\),
\[
  \|\F'(v)h\|_{H^{-1}}
  \le \|Ah\|_{H^{-1}} + \|3v^2h\|_{H^{-1}}.
\]
By definition of the norms, \(\|Ah\|_{H^{-1}} = \|h\|_\V = 1\).
Arguing as in Step~3, we obtain 
\[
   \|3v^2 h\|_{H^{-1}}
  \;\le\; \frac{3}{4\pi^2}\,\|v\|_\V^2\,\|h\|_\V
  \;\le\; \frac{3}{4\pi^2}\,r^2.
\]
Hence,
\[
  \|\F'(v)\|
  = \sup_{\|h\|_\V=1}\|\F'(v)h\|_{H^{-1}}
  \;\le\;
  1 + \frac{3}{4\pi^2}\,r^2
  =: M
\]
for all \(v\in B_r\), which is the \(M\) stated in \Cref{thm:mainPDE}.

Moreover, by coercivity of \(A\) and the positivity of the nonlinearity,
we obtain
\[
  \langle \F'(v)h,h\rangle_{\W,\V}
  = \langle Ah,h\rangle_{\W,\V} + \langle 3v^2h,h\rangle_{\W,\V}
  \;\ge\; \|h\|_\V^2,
\]
so that the smallest singular value of \(\F'(v)\) is bounded below by~1.
Therefore
\[
  \sigma_* := \inf_{v\in B}\sigma_{\min}(\F'(v)) \;\ge\; 1.
\]

Thus (A3) holds with the above \(M\) and \(\sigma_*\ge 1\).

\medskip
\emph{Step 5: Forcing term and choice of \(\bar\theta\).}
Let \(\lambda>0\) be fixed and assume that we use a constant
Tikhonov parameter, i.e.\ \(\lambda_k\equiv\lambda\).
For the learned resolvent \(\widehat R_\lambda\), we define the design error
\[
  \varepsilon_\lambda
  := \sup_{v\in B_r}
     \big\|\,\widehat R_\lambda(v)-R_\lambda(v)\,\big\|_{\mathcal L(\mathcal V,\mathcal V)}.
\]
Then, \Cref{thm:main} (Tikhonov--inexact NK) gives, for any iterate
\(v_k\in B_r\),
\[
  \frac{\|\F'(v_k)\delta v_k+\F(v_k)\|_\W}{\|\F(v_k)\|_\W}
  \;\le\;
  \frac{\lambda}{\lambda+\sigma_*^2}
  + M^2\,\varepsilon_\lambda
  =:\theta_k.
\]
Using \(\sigma_*\ge 1\), we further obtain
\[
  \theta_k
  \;\le\;
  \frac{\lambda}{1+\lambda} + M^2\,\varepsilon_\lambda
  =: \bar\theta.
\]
If \(\bar\theta<1\), then the hypothesis
\(\sup_k \theta_k \le \bar\theta<1\) of \Cref{thm:main} is satisfied.

\medskip
\emph{Step 6: Application of \Cref{thm:main} and explicit \(\widetilde L\).}
\Cref{thm:main} then yields that all the inexact
Newton--Kantorovich conclusions hold with modified Lipschitz constant
\[
  \widetilde L
  \;=\;
  \frac{L}{1-\bar\theta}.
\]
Plugging in the PDE-specific values
\[
  L = \frac{3}{2\pi^2}\,r,
  \qquad
  \bar\theta
  = \frac{\lambda}{1+\lambda} + M^2\,\varepsilon_\lambda,
  \qquad
  M = 1 + \frac{3}{4\pi^2}\,r^2,
\]
we obtain the expression stated in \Cref{thm:mainPDE},
\[
  \widetilde L
  \;=\;
  \frac{L}{1-\bar\theta}
  \;=\;
  \frac{\frac{3}{2\pi^2}r}{\,1-\frac{\lambda}{1+\lambda}-M^2\varepsilon_\lambda\,}.
\]

Together with the identification
\(\eta = \|\F'(0)^{-1}\F(0)\|_\V = \|f\|_{H^{-1}}\) from Step~1,
this shows that all conclusions of \Cref{thm:main} hold in the PDE setting with
the explicit constants given in \Cref{thm:mainPDE}.
\end{proof}

\end{document}